%% file: hwa_training.tex
\documentclass[10pt, dvipsnames]{article}

\usepackage{amsmath,amssymb,amsfonts}
\usepackage{hyperref}
\usepackage{graphicx}
\usepackage{xcolor}
\usepackage{booktabs}
\usepackage[left=2cm, right=5cm, top=2cm]{geometry}
\usepackage{acronym}
\usepackage{bm}
\usepackage{authblk}

\DeclareMathOperator{\sign}{sgn}
\DeclareMathOperator{\quantize}{quant}
\DeclareMathOperator{\clipping}{clip}

\title{Hardware-aware training for large-scale and diverse\\
  deep learning inference workloads\\
  using in-memory computing-based
  accelerators}
\author[1]{Malte J. Rasch\footnote{Correspondence: malte.rasch@ibm.com}}
\author[2]{Charles Mackin}
\author[3]{Manuel Le Gallo}
\author[2]{An Chen}
\author[2]{Andrea Fasoli}
\author[3]{Fr\'ed\'eric Odermatt}
\author[1]{Ning Li}
\author[3]{S.R. Nandakumar}
\author[2]{Pritish Narayanan}
\author[2]{Hsinyu Tsai}
\author[2]{Geoffrey W. Burr}
\author[3]{Abu Sebastian}
\author[1]{Vijay Narayanan}

\affil[1]{IBM Research, TJ Watson Research Center, Yorktown Heights, NY~~USA}
\affil[2]{IBM Research Almaden, 650 Harry Road, San Jose, CA~~USA}
\affil[3]{IBM Research Europe, 8803 R\"uschlikon, Switzerland}

\acrodef{DNN}[DNN]{deep neural network}
\acrodef{AI}[AI]{artificial intelligence}
\acrodef{MVM}[MVM]{matrix-vector multiplication}
\acrodef{AIMC}[AIMC]{analog in-memory computing}
\acrodef{NVM}[NVM]{non-volatile memory}
\acrodef{QAT}[QAT]{quantization-aware training}
\acrodef{CNN}[CNN]{convolutional neural network}
\acrodef{ADC}[ADC]{analog-to-digital converter}
\acrodef{DAC}[DAC]{digital-to-analog converter}
\acrodef{HWA}[HWA]{hardware-aware}
\acrodef{RNN}[RNN]{recurrent neural network}
\acrodef{LSTM}[LSTM]{long-short-term memory}
\acrodef{PCM}[PCM]{phase-change memory}
\acrodef{ReRAM}[ReRAM]{resistive random-access memory}
\acrodef{ECRAM}[ECRAM]{electro-chemical random-access memory}
\acrodef{MRAM}[MRAM]{magnetic RAM}
\acrodef{CBRAM}[CBRAM]{conductive-bridge RAM}
\acrodef{FP}[FP]{floating point}
\acrodef{SGD}[SGD]{stochastic gradient descent}
\acrodef{HMM}[HMM]{hidden Markov model}
\acrodef{SWB}[SWB]{Switchboard}
\acrodef{GLUE}[GLUE]{General Language Understanding Evaluation}
\acrodef{FC}[FC]{fully connected}
\acrodef{LR}[LR]{learning rate}
\acrodef{SNR}[SNR]{signal-to-noise ratio}
\acrodef{AIHWKIT}[AIHWKit]{IBM Analog Hardware Acceleration Toolkit}
\acrodef{MRPC}[MRPC]{The Microsoft Research Paraphrase Corpus}
\acrodef{FLOPS}[FLOPS]{Floating Point Operations Per Second}
\acrodef{CMOS}[CMOS]{Complementary Metal-Oxide Semiconductor}

\begin{document}
\maketitle

\newcommand{\todo}[2]{{\color{blue}{TODO ({\color{red}{#1}}\color{blue}): {#2}}}}
\newcommand{\figref}[1]{Fig.~\ref{fig:#1}}
\newcommand{\twofigref}[2]{Figs.~\ref{fig:#1} and \ref{fig:#2}}
\newcommand{\tabref}[1]{Table~\ref{tab:#1}}
\newcommand{\secref}[1]{Sec.~\ref{sec:#1}}
\renewcommand{\eqref}[1]{Eq.~\ref{eq:#1}}
\newcommand{\twoeqref}[2]{Eqs.~\ref{eq:#1} and \ref{eq:#2}}
\newcommand{\bit}{\text{bit}}

\newcommand{\normerror}{\varepsilon}
\newcommand{\normhour}{\normerror^{1h}}
\newcommand{\normsec}{\normerror^{1s}}
\newcommand{\normfp}{\normerror_*}
\newcommand{\normfphour}{\normfp^{1h}}
\newcommand{\normfpday}{\normfp^{1d}}
\newcommand{\normfpyear}{\normfp^{1y}}

\newcommand{\acc}{\mathcal{A}}
\newcommand{\acchour}{\acc^{1h}}
\newcommand{\accsec}{\acc^{1s}}
\newcommand{\accfp}{\acc_*}
\newcommand{\accfphour}{\accfp^{1h}}
\newcommand{\accfpday}{\accfp^{1d}}
\newcommand{\accfpyear}{\accfp^{1y}}

\newcommand{\chanceerror}{\epsilon_\text{chance}}
\newcommand{\testerror}{\epsilon_\text{test}}
\newcommand{\testerrorfp}{\testerror^\text{FP}}
\newcommand{\testhour}{\testerror^{1h}}
\newcommand{\testday}{\testerror^{1d}}
\newcommand{\mvmerror}{\epsilon_M}
\newcommand{\mvmerrorstd}{\epsilon_M^*}

\newcommand{\fnonlin}{f^\text{NL}}
\newcommand{\analog}[1]{\breve{#1}}
\newcommand{\Wana}{\analog{W}}
\newcommand{\wana}{\analog{w}}
\newcommand{\xana}{\analog{x}}
\newcommand{\xxana}{\analog{\mathbf{x}}}
\newcommand{\yana}{\analog{y}}
\newcommand{\yyana}{\analog{\mathbf{y}}}
\newcommand{\yaimc}{\widetilde{y}}
\newcommand{\yyaimc}{\widetilde{\mathbf{y}}}
\newcommand{\yirdrop}{\Delta \yana^{\text{IR-drop}}}
\newcommand{\sigwnoise}{\sigma^\text{w}}
\newcommand{\sigout}{\sigma^\text{out}}
\newcommand{\sigwnoiset}{\tilde{\sigma}^\text{w}}
\newcommand{\inpbound}{c_{\text{input}}}
\newcommand{\fanalogMVM}{\analog{\mathbf{F}}}
\newcommand{\gtarget}{\hat{g}}
\newcommand{\ggtarget}{\hat{\mathbf{g}}}
\newcommand{\gmax}{{\gtarget}_{\text{max}}}
\newcommand{\gmin}{{\gtarget}_{\text{min}}}
\newcommand{\gprog}{g^{\text{P}}}
\newcommand{\gdrift}{g^{\text{D}}}
\newcommand{\gfinal}{\tilde{g}}
\newcommand{\muprog}{\mu^{\text{P}}}
\newcommand{\sigprog}{\sigma^{\text{P}}}
\newcommand{\quant}{\quantize}
\newcommand{\clip}[3]{\clipping_{#1}^{#2}\left(#3\right)}
\newcommand{\round}{\text{round}}
\newcommand{\bout}{b_\text{out}}
\newcommand{\tinf}{t_\text{eval}}
\newcommand{\affinebeta}{\beta}
\newcommand{\affinebetabold}{\bm{\beta}}
\newcommand{\affinealpha}{\gamma}
\newcommand{\affinealphabold}{\bm{\gamma}}
\newcommand{\affinescale}{\affinealpha}
\newcommand{\sinput}{\alpha}
\newcommand{\ymin}{y_\text{min}}
\newcommand{\ymax}{y_\text{max}}
\newcommand{\alphaaws}{c_\text{aws}}
\newcommand{\globalalpha}{\kappa}
\newcommand{\learnglobal}{\tilde{\globalalpha}}
\newcommand{\alphalearn}{\tilde{\gamma}}
\newcommand{\fp}{$\text{FP}_{32}$}
\newcommand{\nm}{{\color{Blue}${}^\diamond$}}
\newcommand{\nmbm}{{\color{Blue}${}^\circ$}}
\newcommand{\std}[1]{{\tiny \color{Gray}{$#1$}}}
\newcommand{\ppl}{{\color{Blue}${}^\ddagger$}}
\newcommand{\idealfirstlast}{{\color{Blue}${}^\dagger$}}
\newcommand{\noIR}{{\color{Blue}${}^\star$}}
\newcommand{\hmm}{{\color{Blue}${}^\square$}}
\renewcommand{\emph}{}

\begin{abstract}
  Analog in-memory computing (AIMC) \acused{AIMC}-- a promising approach for energy-efficient acceleration of  deep learning workloads --  computes \acp{MVM} but only \emph{approximately}, due to nonidealities that often are non-deterministic or nonlinear. 
  This can adversely impact the achievable \ac{DNN} inference accuracy as compared to a conventional \ac{FP} implementation. 
  While retraining has previously been suggested to improve robustness, prior work has explored only a few \ac{DNN} topologies, using disparate and overly simplified \ac{AIMC} hardware models. 
  Here, we use \ac{HWA} training to systematically examine the accuracy of \ac{AIMC} for multiple common \ac{AI} workloads across multiple \ac{DNN} topologies,  and investigate sensitivity and robustness to a broad set of nonidealities. 
  By introducing a new and highly realistic \ac{AIMC} crossbar-model, we improve significantly on earlier retraining approaches. 
  We show that many large-scale \acp{DNN} of various topologies, including \acp{CNN}, \acp{RNN}, and transformers, can in fact be successfully re-trained to show iso-accuracy on \ac{AIMC}. 
  Our results further suggest that \ac{AIMC} nonidealities that add noise to the inputs or outputs, not the weights, have the largest impact on \ac{DNN} accuracy, and that \acp{RNN} are particularly robust to all nonidealities.
\end{abstract}

\section{Introduction}

The ever-increasing compute needed to train and use \acp{DNN}~\cite{sevilla2022compute} have made hardware latency and energy-efficiency a growing concern.  
However, conventional processor architectures (e.g. CPUs, GPUs, etc.) incessantly transfer data between memory and processing through the `von Neumann bottleneck,' inducing time and energy overheads that significantly degrade latency and energy-efficiency.
Numerous hardware concepts have been introduced to accelerate \ac{DNN} training and/or inference~\cite{Sze2017,Verma2020,2021reutherHPEC}, by approximating \acp{MVM} and other arithmetic
with custom floating-point representations such as bfloat16~\cite{bfloat16} or DLFloat~\cite{DLFloat16}, or with
reduced-precision fixed-point arithmetic to quantize synaptic weights and activations~\cite{Sun2020,choi2018pact,  courbariaux2016binarized, rastegari2016xnornet}. 
Model compression and sparsification techniques can further reduce compute requirements by pruning weights and/or activations~\cite{Albericio2016, han2016deep}.

\ac{AIMC} using \ac{NVM} elements is a promising mixed-signal approach for \ac{DNN} acceleration~\cite{Burr2017a,Y2020sebastianNatNano,burr2021ohm}, with weights stored using crossbar arrays of tuneable conductive elements. 
This enables approximate \ac{MVM} computation directly in-memory, by applying activation vectors (as voltages or pulse durations) to the crossbar array, and then reading out analog physical quantities (instantaneous current or accumulated charge)~\cite{merrikh2017high, chang2019,2020murmannTVLSI}.
As a `non-von Neumann' architecture, \ac{AIMC} performs \ac{MVM} operations at the location of the stored weights,  in a highly-parallel, fast, and energy-efficient manner~\cite{chang2019} -- \emph{but only approximately.}

\begin{figure}[tb]
  \centering
  \includegraphics[width=0.75\textwidth]{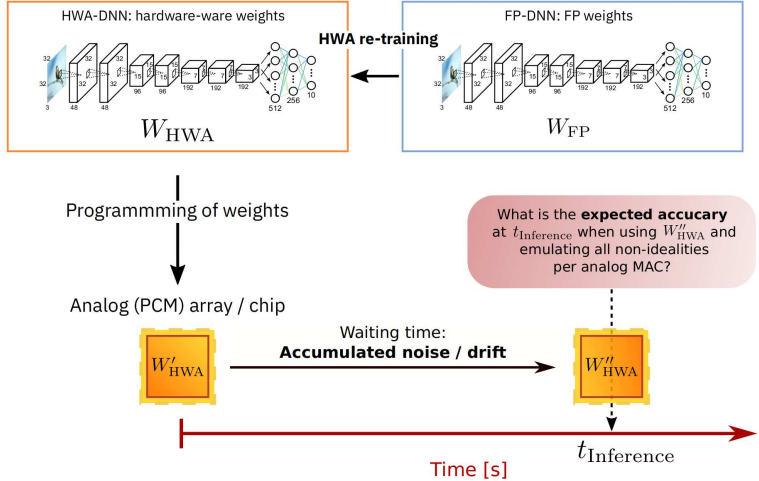}
  \caption{\small 
    In our \acf{HWA} training setup, \acp{DNN} are first trained in $32\bit$ floating-point (\fp), then re-trained in a hardware-aware manner, by adding nonidealities and noise sources into the forward path and using \ac{SGD} to improve the robustness to such generic nonidealities. 
    \ac{HWA} training is only performed once --- no specific device or chip characteristics, such as failure maps, are taken into account during \ac{HWA} training, so resulting models remain widely-deployable. 
    This \ac{HWA}-trained model is then programmed onto \ac{AIMC} multiple times (here in simulation) and \ac{DNN} accuracy is evaluated over time, taking into account conductance drift of \ac{PCM} devices and read noise accumulation~\cite{Nandakumar2019}}.
  \label{fig:inference_scheme}
\end{figure}

The success of reduced-precision digital accelerators proved that \acp{DNN} can tolerate surprisingly coarse
approximations of the underlying \acp{MVM}. 
While naive direct weight-quantization invariably leads to \ac{DNN} accuracy loss, original model accuracies can often be recovered when \acp{DNN} are \emph{re-trained in a quantization-aware manner}, even for aggressive reductions in precision.
Weight-quantization into as few as 2-4~bits can often be tolerated without significant accuracy reduction~\cite{krishnamoorthi2018quantizing,choi2018pact,nagel2021white}. 
This observation led to the development of \ac{QAT} methods, now commonly used when deploying \acp{DNN} onto reduced-precision digital hardware~(e.g.~\cite{Agrawal2021}).

In general, since reducing \ac{MVM} precision decreases the representational power of each \ac{DNN} layer
as compared to a full \ac{FP} implementation, performance naturally suffers once the function approximation becomes too coarse for the task at hand~\cite{nagel2021white}. 
In practice, \ac{QAT} is known to have limits, and \ac{MVM} minimum-precision requirements vary across each \ac{DNN} topology. 
For instance, the first and last layers of \acp{CNN} are invariably implemented with high precision \ac{FP}, even in studies claiming \ac{CNN} iso-accuracy at very low fixed-point precision~\cite{Sun2020, choi2018pact}.

Up to now, it has been unclear how and to what degree \acp{DNN} can be re-trained to maintain accuracy on emerging \ac{AIMC} technology. 
The successes of \ac{QAT} cannot be directly translated onto \ac{AIMC}, since the \ac{MVM} approximations arise from fundamentally different concepts. 
In \ac{AIMC}, weights are represented by conductances that are physical properties of \ac{NVM} devices. 
In many materials, such as \ac{PCM}~\cite{Burr2016,2020legalloJPD}, \ac{ReRAM}~\cite{Jang2015, Jang2014}, \ac{CBRAM}~\cite{Lim2018, Fuller2019}, or \ac{ECRAM}~\cite{li2018capacitor, onen2022nanosecond}, these conductances are effectively continuous physical quantities, and stored weights are \emph{not quantized.} 

That said, effective \ac{AIMC} weight-precision is impacted by various nonidealities, including thermal and 1/f noise, randomization during physical switching induced by electrical and thermal fluctuations, material inhomogenities~\cite{chen2017}, and device-to-device variability introduced during device fabrication or operation.
These issues cause both \ac{MVM} read-out~\cite{Nandakumar2019} and the writing or programming of the conductances~\cite{Papandreou2011, tsai2019inference, Mackin2019} to be erroneous and \emph{non-deterministic}. 
Worse yet, conductances can evolve over time after programming~\cite{Boniardi2010,
  Ambrogio2019, Bruce2021}. 
Finally, any nonlinearities within the analog circuitry performing summation will further degrade \ac{MVM} precision. 
Such imperfections include `IR-drop' voltages on wires and transistors, restrictions on input (output) dynamic-range imposed by discretization and saturation of the \ac{DAC} (\ac{ADC}) components, and random noise or variability in the circuitry.

Whereas \ac{QAT} gets challenging as precision is deterministically reduced,  
\ac{MVM} approximation in \ac{AIMC} is tied to non-deterministic \emph{signal-to-noise ratio}.
A number of previous studies have shown that \emph{noise-aware training} -- simple injection of noise onto weights or
activations during \ac{DNN} training -- can make \acp{DNN} more robust for \ac{AIMC} deployment ~\cite{tsai2019inference, Joshi2020,  yang2022tolerating, gokmen2019iedm, Kariyappa2021, Spoon2021}.
However, such studies have typically been limited to one or two exemplary \acp{DNN} of a particular type (e.g., \acp{CNN}) using only a limited subset of nonidealities such as \ac{NVM} noise.  Other critical \ac{AIMC} system aspects such as output noise, saturation, and circuit nonlinearities have been neglected.  
Moreover, since each study makes different hardware and \ac{NVM}-device choices, it is difficult to generalize, compare, or combine them. Thus more realistic and \emph{standardized} \ac{AIMC} crossbar-models -- which can support comparison of \ac{AIMC} accuracy for \emph{hardware-aware} trained \ac{DNN} models across studies -- are needed.

Although some promising, small-sized \ac{DNN} prototype demonstrations exist~\cite{wan2022compute, khaddam2021hermes, xue2021cmos, fick2022analog, narayanan2021vlsi, Ambrogio2018, yao2020fully}, it remains unclear how robust the \ac{AIMC} deployment of realistically-sized AI workloads will be. 
How will the various nonidealities of \ac{AIMC} hardware impact the \ac{DNN} accuracy, across all the various topographies and thus application domains?  And how much of the lost accuracy could be recovered by \emph{hardware-aware training}?  
Which crossbar-array design choices will be most effective in maintaining accuracy?  And if necessary, what degree of improved device-to-device uniformity might be required -- through better \ac{NVM} device-fabrication -- in order for \ac{AIMC} to succeed on all \ac{DNN} models?
A systematic study comparing the various \ac{DNN} topographies in terms of robustness to \ac{AIMC} nonidealities is needed.

In this paper, we establish a robust \ac{HWA} framework by extending and improving existing training methods for \ac{AIMC} to include previously neglected nonidealities (see \figref{inference_scheme} for an illustration).  
We define a standard inference model for \ac{PCM}-based \ac{AIMC} that can readily be extended to other types of \ac{NVM} devices. 
We explore the functional suitability of \ac{AIMC} across application domains by assessing the robustness of a wide set of \ac{DNN} topographies. 
Finally, we estimate the individual impact of various \ac{AIMC} nonidealities and gauge their relative importance for consideration in future hardware designs.
Functions for our standard evaluation process are provided in an open-source \ac{AIHWKIT}~\cite{Rasch2021AFA}, enabling future studies on noise robustness for \ac{AIMC} to build seamlessly upon our work. 

We find that various \acp{DNN} and AI workloads -- ranging across image classification using \acp{CNN}, text-prediction and speech-to-text conversion using \acp{RNN}, and natural language processing using transformer networks -- can actually be robustly deployed on \ac{AIMC} given proper \ac{HWA} training.  
We show -- for the first time -- iso-accuracy inference results (within 1\% of the \ac{FP} reference) using hardware-calibrated \ac{PCM} models, for five out of the eleven AI workloads tested, even after 1 hour (or more) of conductance drift. 

However, precision requirements are heterogeneous, and not all architectures reach this iso-accuracy target easily even after extensive \ac{HWA} training, pinpointing the need for continued device improvement. 
We find that \acp{CNN} are typically much less robust to various nonidealities and design choices of \ac{AIMC} hardware.
Interestingly, \acp{RNN} -- already well-suited for \ac{AIMC} given their large, dense  \acp{MVM}~\cite{jain2023heterogeneous} -- also seem to be the most robust to the finite \ac{SNR} of \ac{AIMC} hardware.
We further show that among various nonidealities tested, the sensitivity to additive system noise at the output of each crossbar-array is the most critical for achieving good accuracy.

\begin{figure}[tb]
  \centering
  \vglue -1mm
  \includegraphics[width=0.75\textwidth]{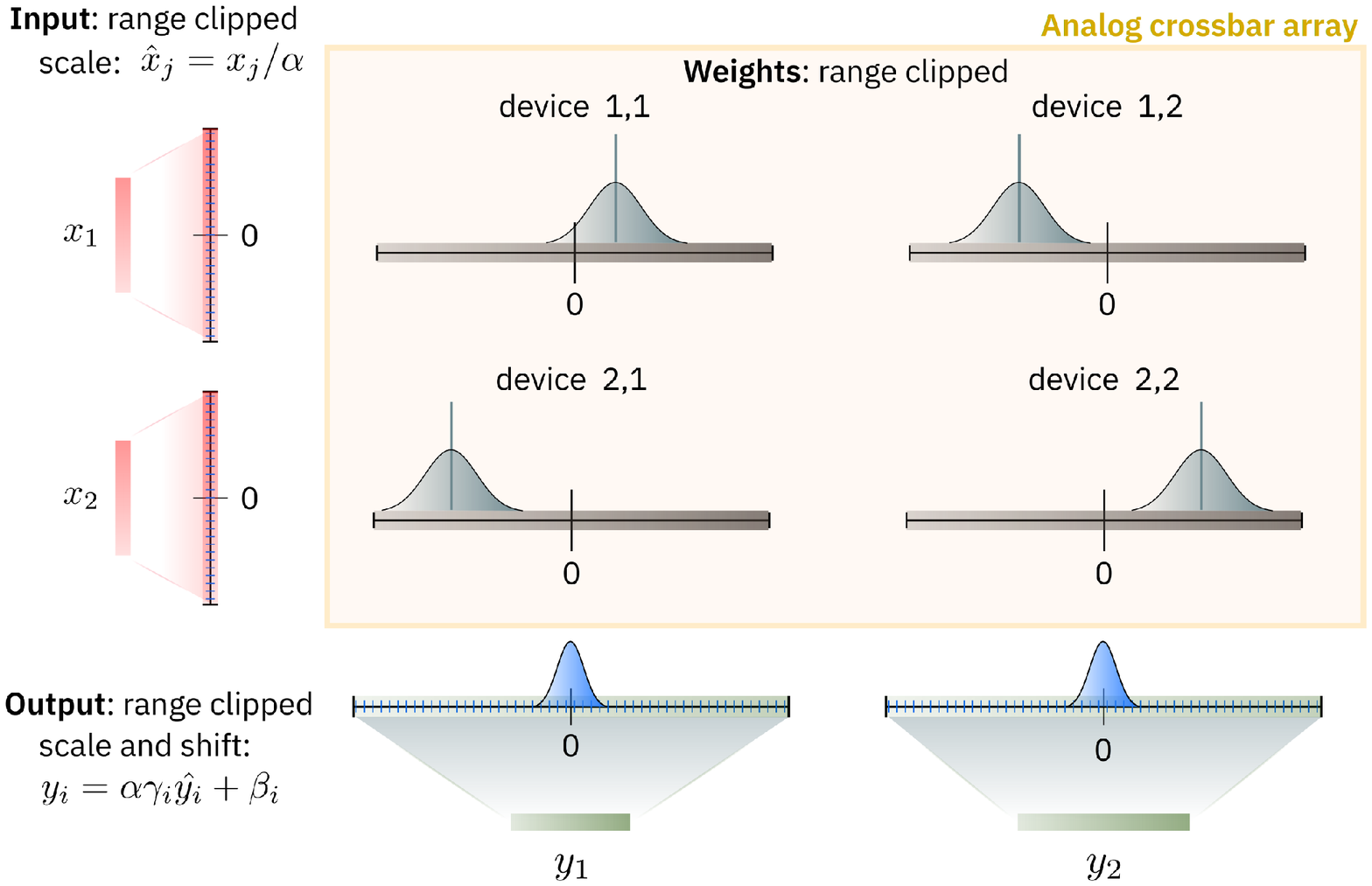}
  \vglue -2mm
  \caption{\small Our \ac{AIMC} crossbar-model assumes that each array or ``tile'' approximates \acf{MVM} $\yyaimc \approx W\mathbf{x}$, using a scalar $\sinput$ (and vectors $\bm{\affinealpha}$ and $\bm{\affinebeta}$) to scale into (from) quantized inputs (outputs) so that input, weight, and output ranges can remain fixed. Negative weights are programmed onto a different conductance for current subtraction, and output noise is fully represented (see \twoeqref{periphery}{analog-mvm}). 
 }
  \label{fig:aimc-mvm-model-illu}
\end{figure}

\section{Results}
\label{sec:results}

\subsection{Analog IMC standard MVM model}

\begin{figure}[t]
  \centering
  \includegraphics[width=1.0\textwidth, clip, trim=3cm 5mm 2cm 5mm]{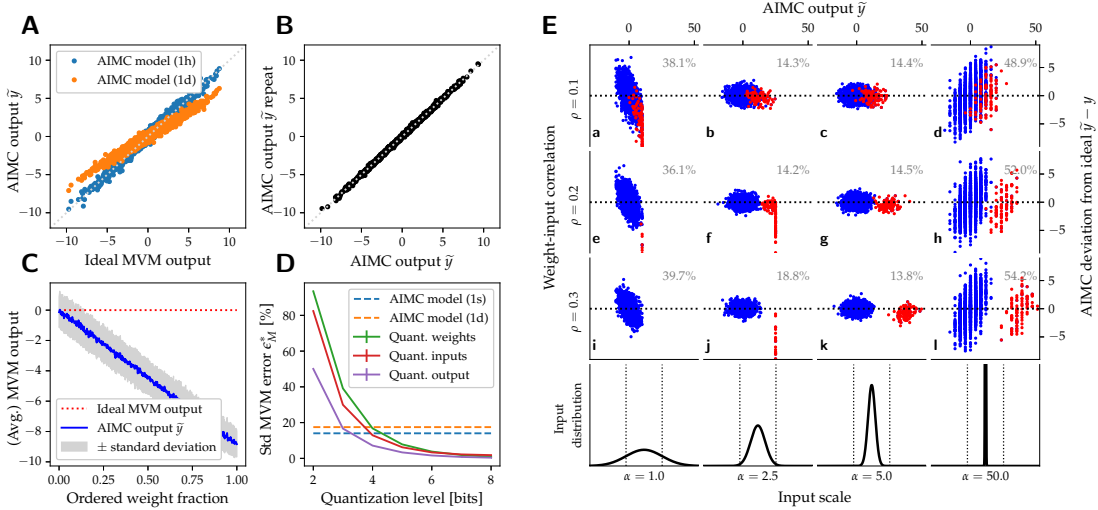}
  \caption{\small 
    (A) Correlations between \ac{AIMC} outputs -- for \acp{MVM} performed between Gaussian random weight matrices and uniform random inputs -- and the ideal (\fp) expected results reveal significant deviations, due primarily to
     weight-programming errors and \ac{PCM} conductance drift (shown here without any mean-drift compensation~\cite{Ambrogio2019}).
    (B) Short-term noise sources induce cycle-to-cycle noise for repeated \ac{MVM} calculations even with the same programmed weight matrix. 
    (C) ``IR-drops'' due to finite wire resistance result from input-position dependency of the accumulated \ac{AIMC} column-currents, and can cause outputs to further deviate from the ideal \ac{MVM}. 
    In an extreme case, an expected 0 output -- the correct result when a linearly-graded weight matrix (ranging from -1 to 1 in order) is read with a constant input on all rows 
    -- can actually deviate drastically due to these position-dependent IR-drops. 
    The deviation induced by these IR-drops increases as the fraction of ordered weights increases from 0 (completely unordered, typical case) to 1 (fully-ordered weights from -1 to 1, extreme case). 
    (D) The \ac{MVM}-error $\mvmerrorstd$ (\eqref{mvm-error}) of our standard \ac{PCM}-based \ac{AIMC} inference model (using the nonideal \ac{MVM} of \twoeqref{periphery}{analog-mvm} together with hardware-calibrated \ac{PCM} conductance noise and drift Eqs.\ref{eq:prog_noise}-\ref{eq:short-term-wnoise}; dotted lines)  $\approx$15\% roughly corresponds to fixed-point quantized digital (solid lines) at $\sim 4$~bits.
   (E) Correlations of \ac{MVM} deviation, $\tilde{y_i}-y_i$, vs.\ desired  \ac{MVM} output $y_i = w_{ij}x_j$ illustrate the importance of proper input scaling $\sinput$, for $w_{ij}\sim {\cal N}(0, 0.246)$ and $x_j\sim {\cal N}(0, 1)$.  
   Red dots mimic the weight-to-activation correlations that \ac{SGD} learning will produce, using $\tilde{x}_j = \rho w_{kj} + (1-\rho) x_j$. 
   Low $\sinput=1$ leads to input clipping (panels a, e, i) and $\mvmerrorstd$ exceeding 35\% (gray text). 
   Intermediate $\sinput$ values can still lead to saturated outputs for correlated inputs, even without input clipping (panels f,j); excessive $\sinput$ values reduce clipping but increase $\mvmerrorstd$ dramatically (panels d, h, l).  
   We optimize $\sinput$ during \ac{HWA} training, then keep it fixed during \ac{AIMC} inference, 
   minimizing $\mvmerrorstd$ regardless of input correlation (panels c, g, k).
 }
  \label{fig:aimc-mvm-model}
\end{figure}

Our standard \ac{AIMC} crossbar-model (see \twofigref{aimc-mvm-model-illu}{aimc-mvm-model}, and \twoeqref{periphery}{analog-mvm} in Methods) encapsulates the critical nonidealities incurred during \ac{MVM} operations, including the fixed dynamic-ranges of physical inputs (limited by maximum pulse-duration), weights (limited by maximum conductance), and outputs (limited by maximum output current). 
Floating point inputs $x$ are mapped to the fixed input range by a global scalar, $\sinput$, which is optimized for each crossbar during \ac{HWA} training, then held fixed for inference.  Such optimization avoids issues created if $\sinput$ is chosen poorly (\figref{aimc-mvm-model}E).
Similarly optimized scales ($\affinealpha_i$) and offsets ($\affinebeta_i$) map \ac{ADC}-counts of each output column to \ac{MVM}-outputs $\yaimc_i$ (see \eqref{periphery}) that can be passed to subsequent digital compute for auxiliary operations (activation functions, etc.)~\cite{jain2023heterogeneous}. 
We further assume a number of nonidealities, such as \ac{PCM} programming errors and drift(\figref{aimc-mvm-model}A), IR-drops within the array (\figref{aimc-mvm-model}C), weight read noise, and system noise (\figref{aimc-mvm-model}B and \eqref{analog-mvm}). 
Each of these parameters has been carefully calibrated to existing \ac{PCM} hardware~\cite{Nandakumar2019}. All parameter settings of the \ac{AIMC} crossbar-model are summarized in \tabref{aimc-parameter}.

We quantify \ac{MVM} errors in computing $\yyaimc$ with respect to the ideal outcome $\mathbf{y}$
through $\mvmerror$, 
the ratio of the $l_2$-norm of the deviation ($\mathbf{y}-\yyaimc$) relative to the $l_2$-norm of the ideal outcome $\mathbf{y}$ (see \eqref{mvm-error}). 
Figure~\ref{fig:aimc-mvm-model}B shows that, even after including \ac{PCM} drift, the effective \ac{MVM} error of the standard \ac{AIMC} crossbar-model we will use throughout the paper roughly corresponds to $4\bit$ fixed-point quantization of weights or inputs.

\subsection{DNN accuracy impact  when directly  using  AIMC }

To test the effect of \ac{AIMC} nonidealities on a variety of AI workloads, we consider 11 medium- to large-scale \acp{DNN} of various topologies as a benchmark set (see \tabref{dnn-properties}). These cover a wide spectrum of target-applications (image classification, natural language processing, speech-to-text), network topologies (convolutional, recurrent, transformer with attention), model sizes (from 0.3M to 108M parameters), crossbar utilization (from 4\% to 86\%), total number of \acp{MVM} per data input (from 2.2K to 240K), \ac{MVM} sizes (from 0.1G to 96.5G flops), average weight-matrix reuse-factor per data-input (from 17 to 1285), and network depth (up to 121 layers). Our benchmark set thus covers a wide variety of network topologies and challenges for \ac{AIMC}.

\begin{table}
  \centering

  {\small
  \begin{tabular}[l]{lc|ccccccc}
    \toprule
    DNN& Type &\# par.&\# mapped&\# tiles&util.&\# MVM& $\langle$reuse$\rangle$&flops\\
    \midrule
    ResNet-32 {\tiny CF10} & \textbf{C} &0.36M &0.36M & 34 & 4.0\%  & 14K & 435. & 0.1G\\
    WideRN-16 {\tiny CF100}   &\textbf{C}  & 11M &11M   & 68 & 61.7\% & 19K & 286. & 3.1G\\
    ResNet-18\idealfirstlast {\tiny ImNet} & \textbf{C} & 11.7M &11.2M & 75 & 56.8\% & 40K &533. & 3.4G\\
    ResNet-50\idealfirstlast {\tiny ImNet} & \textbf{C} & 25.6M &23.4M  & 149 & 60.0\% & 74K  & 497. & 7.9G\\
    DenseNet-121\idealfirstlast {\tiny ImNet} & \textbf{C} & 8M &6.9M & 266 & 9.8\% & 143K & 537. &  5.4G\\
    WideRN-50\idealfirstlast {\tiny ImNet} & \textbf{C}& 69M &66.8M  & 296 & 86.0\% & 101K & 342. & 22.6G\\
    BERT-base {\tiny MRPC} & \textbf{T}      & 108M &85M  & 486 & 67.1\% & 61K & 126. & 21.8G \\
    Albert-base {\tiny MRPC} & \textbf{T} & 12M &7.8M & 48  & 61.7\% & 61K & 1285. & 21.8G \\
    Speech\hmm \tiny{SWB300} & \textbf{L}    & 30M &30M   & 153 & 74.8\% & 2.5K & 17. & 0.9G\\
    LSTM {\tiny PTB} & \textbf{L}            & 19.8M &13.3M & 88 & 57.5\% & 2.2K & 26. & 0.7G\\
    RNN-T & \textbf{L}                       & 57M &57M & 304 & 71.6\% & 240K & 790.$^{*}$ & 96.5G\\
    \bottomrule
  \end{tabular}
}
\caption{\small { 
A wide range of \acp{DNN} topologies (Type) and sizes are studied, including \acp{CNN} (\textbf{C}), \acp{LSTM} (\textbf{L}), and Transformers (\textbf{T}).  
    For each \ac{DNN}-model and dataset, model size is quantified by 
    number of parameters (\# par.);
    number of parameters mapped to analog crossbars (\# mapped); 
    number of 512 $\times$ 512 crossbars (\# tiles) needed for naive mapping (each weight matrix gets at least 1 tile); overall utilization of the devices within the tiles (util.); 
    total number of \acp{MVM} per input data (\# \ac{MVM}); 
    average tile-reuse for one input data ($\langle$reuse$\rangle$; ${}^*$ for maximal input length in dataset); 
    and the number of \fp\ operations in the mapped \ac{MVM} for one input data (\ac{FLOPS}).
    \idealfirstlast first conv-layer and last \ac{FC} layer in $\text{FP}_{32}$, \hmm 
    additional hidden-Markov-model used as decoder.}
 \label{tab:dnn-properties}
}
\end{table}

For comparison, we first directly map weights produced by standard \ac{SGD}-training in \fp\ onto our standard \ac{AIMC} crossbar-model and evaluate the resulting test error,
to measure the accuracy drop (with respect to the \fp\ model) due to all the various \ac{AIMC} nonidealities. 
For each crossbar, we first clip raw weight-values to 2.5$\times$ the standard deviation of the weight distribution, to disregard outliers and compactify the weight distribution. 
Output scales $\affinescale_i$ are initially estimated according to the absolute maximum value for each column (see
\eqref{initial-mapping}). To adjust the digital parameters of our standard \ac{AIMC} crossbar-model for these directly-mapped-from-software weights, we use our \ac{HWA}-training flow -- but without any weight noise injection, with weight learning rates set to zero, and for only 1000 batches.   
As expected, such ``direct'' mapping of \acp{DNN} onto \ac{AIMC}, without any additional retraining of the weights, generally results in significant increases in test error (accuracy drop) in comparison to the floating point
reference (\tabref{dnn-direct}). 

Direct comparison of accuracy values between \acp{DNN} is complicated by the fact that these various AI tasks exhibit different worst-case (random guessing) and best-case (well-trained \ac{DNN} model) accuracies.
To quantify and compare accuracy drop across different topologies, we therefore define a \emph{normalized} relative accuracy $\accfphour$,
which re-scales the \ac{AIMC} test error $\testhour$ (at 1 hour \ac{PCM} drift) by the distance between the original \fp\ test error and the ``chance'' test error from random guessing, as follows:
\begin{equation}
\accfphour = 1 - \frac{\testhour - \testerrorfp}{\chanceerror - \testerrorfp}.
\label{eq:accfphour}
\end{equation}
Thus a value of $\accfphour=100$\% means that the \ac{AIMC} \ac{DNN} achieves the \emph{same} accuracy as the \fp\ reference model (no accuracy drop at all), while a value of
$\accfphour=0$\% implies that the \ac{AIMC} crossbar-model is so inaccurate that it is indistinguishable from random guessing.

Ideally, deploying a given \ac{DNN} in an \ac{AIMC} system should have \emph{no impact on model accuracy}. 
We define our iso-accuracy target as $\accfphour>99$\%, allowing less than a $1$\% drop in accuracy, as judged
relative to the distance between the \fp\ reference accuracy and the chance (random guessing) accuracy-floor.
\tabref{dnn-direct} shows that direct \ac{AIMC} mapping fails to achieve this iso-accuracy target for almost all of the
\acp{DNN} tested, establishing both the \emph{challenge} posed by the nonidealities existing in \ac{AIMC} (as compactly encapsulated
by our standard crossbar-model, \twofigref{aimc-mvm-model-illu}{aimc-mvm-model}), as well as the
\emph{need} for \ac{HWA} training methods that can greatly improve the robustness and reduce these accuracy drops.

\begin{table}
  \centering
  {\small
  \begin{tabular}[l]{l|c|cc|cc}
    {\bf Direct mapping}& \multicolumn{3}{l|}{Test Error in \%} & \multicolumn{2}{l}{Normalized accuracy} \\
    \toprule
    DNN & \fp & 1~hour &  1~year & $\accfphour$ [\%] & $\accfpyear$ [\%] \\
    \midrule
    ResNet-32 {\tiny CF10} &5.80 & 11.68 \std{0.16} &15.21 \std{0.40} &93.0&88.8\\
    WideRN-16 {\tiny CF100} &20.00 & 26.02 \std{0.10} &29.42 \std{0.28} &92.4&88.1\\
    ResNet-18\idealfirstlast {\tiny ImNet} &30.50 & 47.91 \std{0.21} &55.00 \std{0.41} &74.9&64.7\\
    ResNet-50\idealfirstlast {\tiny ImNet} &23.87 & 32.51 \std{0.09} &38.18 \std{0.31} &88.6&81.2\\
    DenseNet-121\idealfirstlast {\tiny ImNet} &25.57 & 47.38 \std{0.24} &59.81 \std{0.71} &70.7&53.9\\
    WideRN-50\idealfirstlast {\tiny ImNet} &21.53 & 40.07 \std{0.13} &45.45 \std{0.30} &76.3&69.5\\
    BERT-base {\tiny MRPC} &14.60 & 15.99 \std{0.23} &18.36 \std{0.30} &97.3&92.7\\
    Albert-base {\tiny MRPC} &15.08 & 31.50 \std{0.10} &31.62 \std{0.03} &67.8&67.5\\
    Speech\hmm \tiny{SWB300} &14.05 & 16.11 \std{0.02} &16.32 \std{0.02} &97.6&97.4\\
    LSTM {\tiny PTB} &72.90 & 73.16 \std{0.01} &73.28 \std{0.01} &$\boldsymbol{99.0}$&98.6\\
    RNN-T {\tiny SWB300} &11.80 & 28.16 \std{0.10} &29.08 \std{0.16} &81.4&80.4\\
    \bottomrule
  \end{tabular}
}
\caption{\small Inference results using \ac{AIMC} for the 11 benchmark
  \acp{DNN} when deployed directly without any weight retraining.  
  Test errors in \% $\pm$ standard error of mean (across 24 inference repeats)
  are shown after 1 hour and 1 year of \ac{PCM} drift (center columns) and compared to the
  original \fp\ test error (leftmost column).  Digital parameters needed for the \ac{AIMC}
  crossbar-model are estimated by training briefly with the \ac{AIMC} \ac{MVM} in the forward pass (1000 batches), but without touching the directly-mapped analog weights.  This helps adjust statistics of each batch norm to the new output distributions caused by the \ac{AIMC} \acp{MVM}. 
  During these 1000 batches, we estimate $\sinput$ by averaging the maximal absolute inputs for each batch during the first 500 batches, and then allow \ac{SGD} to tune it further during the second half of the brief digital-parameter-only \ac{HWA}-training.
  Righthand columns show normalized accuracy values after 1 hour ($\accfphour$) 
  and 1 year ($\accfpyear$), as scaled to the range between the \fp\ reference test-error
  and the test-error obtained by random guessing. Nearly all models fail to achieve
  \emph{iso-accuracy}, as defined by  $>99$\% in this normalized accuracy and indicated in bold font.  Note that for BERT and Albert only one \ac{GLUE} task (\ac{MRPC}) is used here. 
}
 \label{tab:dnn-direct}
\end{table}

\subsection{HWA training improves AIMC accuracy for all DNNs}

Building on previous approaches (e.g.~\cite{Joshi2020, Kariyappa2021, gokmen2019iedm}), we set out to retrain these 11 \acp{DNN} in a \acf{HWA} manner. 
In our methodology for \ac{HWA} training followed by delayed inference (\figref{inference_scheme}), each \ac{DNN} is retrained with noise injection using \ac{SGD}. 
But in contrast to earlier approaches, we incorporate a much more comprehensive and realistic set of software-simulated \ac{AIMC} nonidealities, including dynamic-range limitations, weight-programming errors, \ac{PCM} drift and system noise. 
Once a given \ac{DNN} is trained and mapped to \ac{AIMC}, inference is then gauged for noise and drift
at various delays (1 second, 1 hour, 1 day, and 1 year) after programming the weights into the crossbar arrays. 
We also introduce a set of \ac{AIMC} characteristics including input, output and weight-scales (see \figref{aimc-mvm-model} and Methods),
and introduce a new approach for optimizing these scaling-factors during \ac{HWA} training for use during inference (see \secref{input-range-learning}).

As shown in \tabref{hwa-results}, our \ac{HWA} training approach significantly improves achievable accuracy for \ac{AIMC} across the full set of benchmark \acp{DNN} results.  
The normalized accuracies (relative to the \fp\ model) at one hour after programming are all higher than 96\% ($\accfphour$, towards right edge of \tabref{hwa-results}).  This represents a  significant improvement over  `direct' weight mapping without retraining
shown earlier (\tabref{dnn-direct}), while establishing a new state-of-the-art in \ac{HWA} training, as revealed by detailed comparisons on  ResNet-32 with CIFAR10 (see \tabref{hwa-validation}).

\tabref{hwa-results} indicates that five out of the 11 AI workloads can be trained to reach the $\accfphour> 99$\% iso-accuracy target, including the BERT transformer model as well as all workloads based on \acp{LSTM} (last 3 rows, see `Type' column in \tabref{dnn-properties}). 
Most of the remaining workloads use \ac{CNN}s and exhibit more-pronounced accuracy drops of up to $3.6\%$ on \ac{AIMC}, although one \ac{CNN} does reach iso-accuracy (WideResNet-32 on Cifar100).

For some \acp{DNN}, we find that the regularization effect of the added \ac{AIMC} nonidealities allows \ac{HWA}-training to actually improve the attainable accuracy (compare test-errors at 1 sec after programming for WideRN-16 and BERT).
Both \acp{RNN} and transformers are quite robust when subject to \ac{PCM} conductance drift over longer periods as well.  
The rightmost column of \tabref{hwa-results} shows the long-term relative accuracy of the \acp{DNN}, $\accfpyear$, for an hypothetical 1 year after programming without weight refresh.

While the \acp{RNN} and transformers remain near iso-accuracy over time, larger \acp{CNN} with higher resolution ImageNet inputs show the largest drop in accuracy. 
The deep DenseNet-121 (121 layers), as well as the large WideResNet-50 (69M parameters) models are clearly the most challenging for \ac{AIMC}.
That said, the resiliency to long-term drift is greatly improved by \ac{HWA}-training as compared to ``direct'' deployment without retraining. 
For instance, the \ac{HWA}-trained models for both the Speech-SWB300 and LSTM-PTB models remain iso-accurate out to a year,  unlike the directly-mapped models (\tabref{dnn-direct}).

In general, we find that \acp{CNN} are more difficult to train to iso-accuracy for \ac{AIMC} deployment compared to \acp{RNN} and transformers.  
In terms of \ac{AIMC} workload execution latency and system mapping~\cite{jain2023heterogeneous}, 
\acp{CNN} are already less well-suited for resistive crossbar arrays due to the uneven temporal re-use between layers and spatial under-utilization of the large analog tiles by the small kernel matrices (see \tabref{dnn-properties}),
although some optimization and mapping tricks~\cite{rasch2019rapa} are available.
Our results here indicate that \ac{AIMC} noise-robustness issues will pose additional challenges when implementing \acp{CNN} onto \ac{AIMC} systems.

\begin{table}
  \centering
  {\small
  \begin{tabular}[l]{l|c|llll|cc}
   {\bf HWA training}& \multicolumn{5}{l|}{Test Error in \%} & \multicolumn{2}{l}{Normalized accuracy} \\
    \toprule
    DNN & \fp &1 sec & 1 hour & 1 day & 1 year & $\accfphour$ & $\accfpyear$\\
    \midrule
ResNet-32 {\tiny CF10} &5.80 & 6.79 \std{0.02} &7.08 \std{0.03} &7.50 \std{0.04} &8.36 \std{0.06} &98.5&97.0\\
WideRN-16 {\tiny CF100} &20.00 & 19.75 \std{0.02} &20.03 \std{0.02} &20.27 \std{0.02} &21.31 \std{0.04} &$\boldsymbol{100.0}$&98.3\\
ResNet-18\idealfirstlast {\tiny ImNet} &30.50 & 31.57 \std{0.06} &32.21 \std{0.10} &32.94 \std{0.11} &35.31 \std{0.27} &97.5&93.1\\
ResNet-50\idealfirstlast {\tiny ImNet} &23.87 & 24.52 \std{0.02} &24.82 \std{0.05} &25.24 \std{0.06} &27.16 \std{0.13} &98.8&95.7\\
DenseNet-121\idealfirstlast {\tiny ImNet} &25.57 & 27.39 \std{0.07} &28.22 \std{0.10} &29.36 \std{0.15} &36.09 \std{0.31} &96.4&85.8\\
WideRN-50\idealfirstlast {\tiny ImNet} &21.53 & 23.38 \std{0.05} &23.82 \std{0.04} &24.42 \std{0.07} &27.87 \std{0.14} &97.1&91.9\\
BERT-base {\tiny GLUE8} &17.47 & 17.43 \std{0.09} &17.55 \std{0.12} &17.58 \std{0.12} &17.99 \std{0.12} &$\boldsymbol{99.8}$&98.9\\
Albert-base {\tiny GLUE8} &19.46 & 20.52 \std{0.18} &20.45 \std{0.16} &21.08 \std{0.18} &22.18 \std{0.21} &97.8&94.0\\
Speech\hmm \tiny{SWB300} &14.05 & 14.26 \std{0.03} &14.29 \std{0.03} &14.35 \std{0.03} &14.52 \std{0.03} &$\boldsymbol{99.7}$&$\boldsymbol{99.4}$\\
LSTM {\tiny PTB} &72.90 & 72.94 \std{0.01} &72.94 \std{0.01} &72.96 \std{0.01} &73.04 \std{0.01} &$\boldsymbol{99.8}$&$\boldsymbol{99.5}$\\
RNN-T {\tiny SWB300} &11.80 & 12.24 \std{0.15} &12.36 \std{0.09} &12.56 \std{0.13} &13.28 \std{0.28} &$\boldsymbol{99.4}$&98.3\\
    \bottomrule
  \end{tabular}
}
\caption{\small Test error in \% $\pm$ standard error of mean (across 15-25 inference
    repeats per training trial and up to 3 training trails) for \ac{DNN} deployment on \ac{AIMC} crossbars after \ac{HWA} training. 
    Rightmost two columns show the normalized accuracy, scaled between the FP reference and chance error, at 1 hour and 1 year after weight programming.    
    Note that \ac{PCM} drift is a post-programming physical effect that is initially rapid but then slows down logarithmically in time~\cite{2020legalloJPD}. 
    This means that the multiplicative conductance-changes induced by drift between 1 second and 1 hour (time-since-programming increased 3600$\times$), and between 1 hour and 1 year (time-since-programming increased 8760$\times$) are actually quite  similar. 
    \ac{HWA} training hyper-parameters (injected noise strength, etc.) were chosen to produce the best average accuracy across the four widely-spaced time-points shown here.  
    Other choices could be made to focus just on performance in either longer or shorter periods of drift. 
    Models deemed \emph{iso-accurate} ($\accfphour, \accfpyear >$ 99\%) are marked in bold.  
    BERT and Albert results are averaged across eight GLUE tasks, as evaluated on validation datasets; SWB300 results are averaged over two benchmark tasks; results for Speech--SWB300 use \ac{HWA} with distilling (see Methods). 
    }
 \label{tab:hwa-results}
\end{table}

\subsection{Sensitivity of HWA trained models to various AIMC nonidealities}

To determine which nonidealities are particularly problematic for analog inference across \acp{DNN}, we ``stress test'' our \ac{HWA}-trained models. 
For each individual nonideality, such as \ac{PCM} programming-error or IR-drop, we vary its strength and evaluate the resulting inference accuracy across \acp{DNN} using our base \ac{HWA}-trained model.
Our standard \ac{AIMC} \ac{MVM} model exhibits $\mvmerrorstd\approx 15$\% (see \figref{aimc-mvm-model} and \eqref{mvm-error}), but combines many nonidealities. 
To estimate the relative accuracy impact due to each individual nonideality, we boost only that parameter value until \ac{MVM} error increases to $\mvmerrorstd=20$\%, and then re-measure \ac{DNN} accuracy.  

\begin{figure}[t]
  \centering
  \includegraphics[width=\textwidth, clip, trim=0cm 0.0cm 0 0cm]{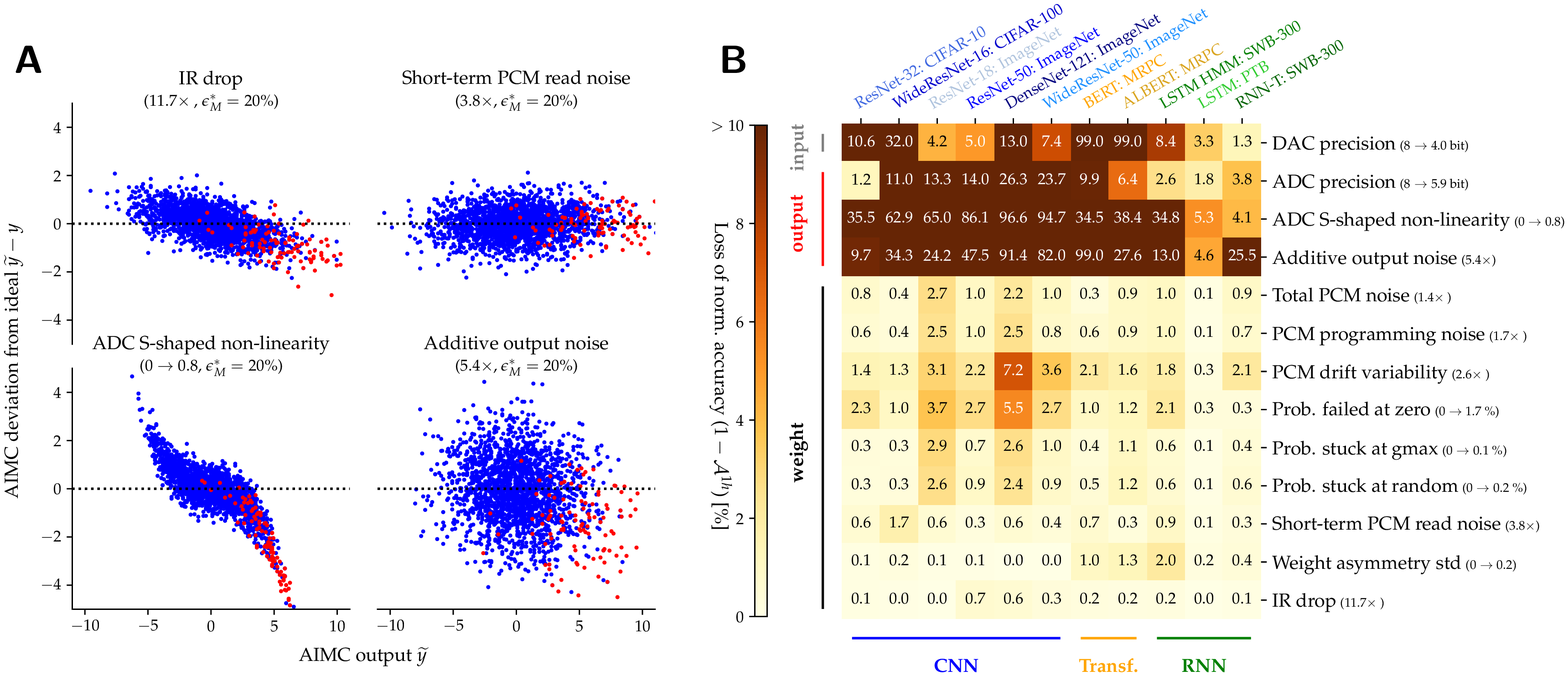}
  \vglue -3mm
  \caption{\small 
     Comparison of the relative impact of various \ac{AIMC} nonidealities on \ac{DNN} accuracy. 
     (A) \ac{AIMC} deviations ($\tilde{y}-y$) from the ideal \ac{MVM} output ($y$) are shown, for uncorrelated (blue dots) and weakly-correlated ($\rho=0.05$) random activations (red dots), as a single nonideality is increased until standard \ac{MVM} error reaches $\mvmerrorstd=20\%$.  
     All other parameters remain fixed to our standard \ac{AIMC} crossbar-model ($\mvmerrorstd=15\%$, \figref{aimc-mvm-model}).
     For instance, IR-drop needs to be scaled 11.7$\times$ to incur $\mvmerrorstd=20\%$. 
     Even at constant $\mvmerrorstd=20\%$, \ac{MVM} deviations are structured differently and thus the impact on \ac{DNN} accuracy can vary significantly.
     (B) Grid shows loss in normalized accuracy ($\acchour$) over the base HWA-trained model at 1 hour after programming when boosting a given nonideality to  $\mvmerrorstd=20\%$.
     Thus 0\% means no accuracy impact despite the amplified nonideality, whereas 100\% means a drop to chance level. For the \ac{HMM} \ac{LSTM} sensitivity is reported for a portion of the training set (instead of the benchmark set) directly on the \ac{LSTM} output without the Hidden-Markov-Model to speed up computations. 
     For the transformer models, only one \ac{GLUE} task is evaluated (\ac{MRPC}). 
    }
  \label{fig:sensitivity}
\end{figure}

Even at constant \ac{MVM} error, each parameter changes a different aspect of the \ac{AIMC} compute.
For instance, output noise is applied at each \ac{MVM}, whereas \ac{PCM} programming errors are only applied during programming and then persist throughout inference. 
Other nonidealities such as IR-drop or ADC ``S-shaped'' nonlinearity change the \emph{shape} of the \ac{MVM} deviations (\figref{sensitivity}A), causing large outputs to incur very significant \ac{MVM} error.  
As a result, even at an identical \emph{average} \ac{MVM} error of $\mvmerrorstd=20$\%, the impact on \ac{DNN} accuracy can be much more pronounced.
Such nonidealities are particularly detrimental for \ac{DNN} inference, and thus deserve additional attention in  future hardware designs or \ac{HWA} training methods.

To gauge the relative impact of each individually-boosted nonideality parameter, \figref{sensitivity}B shows the loss in normalized accuracy ($\acchour$), defined not with respect to the \fp\ model error ($\accfphour$ (\eqref{accfphour})), but with respect to our standard \ac{AIMC} crossbar-model (at 1 hour drift). 
A value of $0$\% means that boosting this particular nonideality has \emph{no impact on accuracy}, as compared to our standard \ac{AIMC} crossbar-model. 
A value of $100$\% means that simply boosting this nonideality to the same \ac{MVM} error of $\mvmerrorstd=20$\% has degraded
\ac{DNN} accuracy to the level of random guessing.

Clearly, \ac{DNN} accuracy reacts vastly differently to individual nonidealities.
We observe that nonidealities that effectively \emph{add noise to the inputs or outputs} -- such as \ac{ADC} and \ac{DAC} resolution, additive output noise, and S-shaped non-linearity of the \ac{ADC} -- have the largest impact on the \ac{DNN} accuracy, as normalized to impact on average \ac{MVM} error. 
\acp{CNN} are the most sensitive \ac{DNN} topology, while \acp{RNN} are the least sensitive (in particular the PTB-LSTM network).

Nonidealities that mostly affect \emph{weight-precision} (all other nonidealities listed in \figref{sensitivity}B), have a much less severe impact on the \ac{DNN} accuracy.  In contrast to additive output noise, such weight-related nonidealities all scale with the input norm, and thus disappear when no inputs are given. 
Since it arises from large currents, IR-drop becomes negligible when either inputs or weights are reduced (in either amplitude or occurrence). Such weight-related nonidealities impact \acp{CNN} slightly more than \acp{RNN} or transformers. 
In particular, DenseNet-121 with small kernel matrices and a high tile re-use factor seems the most affected by weight disturbances. \figref{sensitivity} shows it is not enough to focus only on weight-related nonidealities, as most previous studies have done, when investigating \ac{AIMC}.

We use this sensitivity analysis to assess additional nonidealities which our standard \ac{AIMC} crossbar-model
assumes to be perfect.  
For instance, imperfect device yield -- where some fraction of the weight conductances are ``stuck'' either at zero (\ac{PCM} reset), at $\gmax$ (\ac{PCM} set), or at some intermediate random value -- 
is shown to have the same modest effect on \ac{DNN} accuracy as other weight-related parameters.
Weight asymmetry -- a systematic
difference in conductance for positive versus negative inputs such that
$- w (-|x|) \neq w (|x|$) -- is shown to have only modest impact on \ac{DNN} accuracy.
Interestingly, \acp{RNN} and transformers are the models impacted by such polarity-dependent device response, since the ReLU activations used in \acp{CNN} cannot create negative inputs. 
Finally, \emph{systematic} \ac{PCM} programming errors 
-- applied once to the conductance values and then remaining constant through repeated \acp{MVM} 
-- are shown to have a slightly larger effect than the \emph{cycle-to-cycle} short-term \ac{PCM} read-noise that gets redrawn for every \ac{MVM}.

\subsection{AIMC robustness of DNN topologies}

\begin{figure}[t!]
  \centering
  \includegraphics[width=\textwidth, clip, trim=1.5cm 0.1cm 0 0cm]{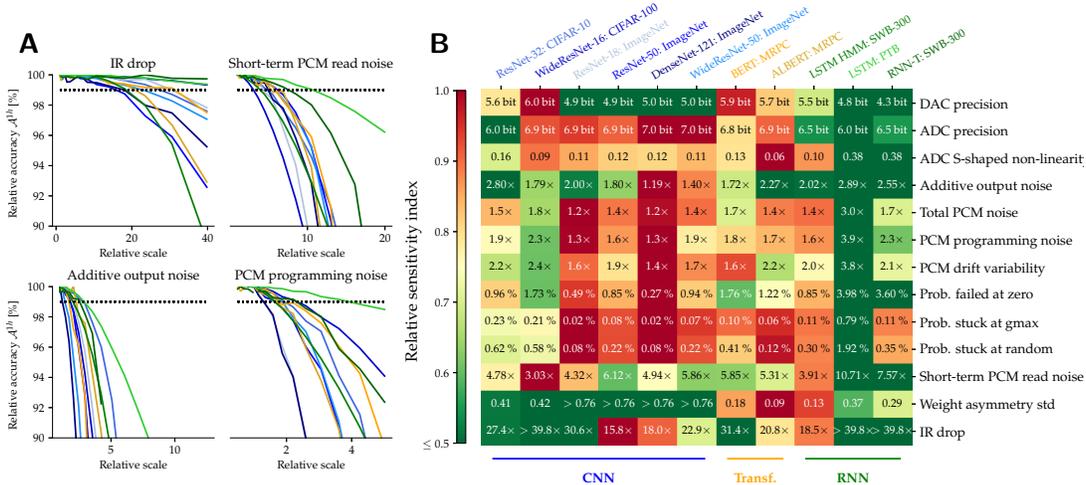}
  \caption{\small Sensitivities of individual \ac{AIMC} nonidealities across \acp{DNN}. 
    (A) As a single nonideality parameter is increased from the our standard setting, accuracy $\acchour$ eventually drops to 99\% (compared to accuracy of the standard \ac{AIMC} crossbar-model).  Four nonidealities are shown, with \ac{DNN} line-colors matching the textlabel color in (B). 
    (B) Grid shows $x^\ast$, the threshold value at which that particular nonideality produces $\acchour=99$\% (\ac{DNN}'s curve crosses dotted-line in (A)).
    For instance, reducing \ac{DAC} precision from $8\bit$ down to $4.8\bit$, while maintaining all other parameters from the standard \ac{AIMC} crossbar-model, causes exactly 1\% additional accuracy loss in the LSTM-PTB model. 
    Text-label colors at top match the lines in (A); grid colors reflect relative sensitivity index $r_S = \frac{x^\ast - \min x^\ast}{\max x^\ast - \min x^\ast}$, with min and max values taken across all \acp{DNN}. 
    $r_s=1$ (red) indicates the most sensitive and $r_s=0$ (green) the least sensitive \ac{DNN}.
    \acp{RNN} are generally observed to be more robust to \ac{AIMC} nonidealities than \acp{CNN}, 
    even with the limited hyper-parameter tuning available for RNN-T due to its large number of \ac{MVM} \ac{FLOPS} and parameters.
    }
  \label{fig:specifications}
\end{figure}

To extract the specific sensitivities of each individual \ac{DNN}, we find the threshold value $x^\ast$ at which each nonideality degrades accuracy to $\acchour(x) = 99$\%, with respect to the standard \ac{AIMC} crossbar-model.  
From scans of $\acchour$ as each nonideality is increased (\figref{specifications}A), we use linear interpolation to identify $x^\ast$ from the intersection with the dotted line at $\acchour = 99$\%.

The grid in \figref{specifications}B shows this threshold value  $x^\ast$, for each nonideality and each \ac{DNN}.
For example, considering just total \ac{PCM} noise, even small increases beyond the current hardware-calibrated values markedly degrade ResNet-18 ($x^\ast = 1.2\times$ for $\acchour=99$\% ), while LSTM-PTB is not affected until this particular nonideality is significantly larger ($x^\ast=3\times$).
The colors ranging from red to green in \figref{specifications} illustrate the relative sensitivity among the
\acp{DNN}, obtained by scaling $x^\ast$ linearly between the minimal and maximal values across the 11 \acp{DNN}. 
For many of these nonidealities, yet again \acp{RNN} tend to be the most robust, followed by small \acp{CNN} on the CIFAR dataset. 

Some nonideality parameters can be increased quite dramatically with respect to our standard \ac{AIMC} crossbar-model baseline. 
For instance, \ac{DAC} precision can be lowered from $8\bit$ to $6\bit$ without any retraining, with little accuracy impact across all \acp{DNN} -- this could produce considerable energy savings and throughput improvement for \ac{AIMC} designs. 
Also, IR-drop can be increased beyond than the baseline before becoming problematic, and short-term weight noise could be up to $3\times$ larger, similarly informing future \ac{AIMC} designs, both with and without \ac{PCM} devices.
While direct examination of \figref{specifications} might suggest that IR-drop could be increased by 15$\times$ without issue, note that the assumptions inherent in our IR-drop calculations, concerning average rather than instantaneous currents, imply a small safety margin of perhaps 3$\times$ (see Methods).

We also estimated the effect of imperfect \ac{PCM} device yield. 
Even the least robust model can tolerate $0.27$\% failed-at-zero devices (stuck in the reset state, at random locations), rising to 3-4\% for some of the \acp{RNN}. 
However, \ac{DNN} accuracies are more sensitive to devices stuck either at random intermediate conductance values or at $\gmax$ (in the set state). 
As few as $0.02$\% such failed devices would already cause a noticeable accuracy drop in some large \acp{CNN}.  
However, our analysis only assumes one pair of conductances per weight --- since many existing \ac{AIMC} designs provide multiple pairs of \ac{PCM} devices per weight~\cite{khaddam2021hermes, narayanan2021vlsi}, such additional redundancy can potentially counteract such stringent device yield requirements.

\subsection{Impact of weight distributions on AIMC MVM fidelity}

The \ac{MVM} error of each \ac{AIMC} tile is affected by the shape of the weight distributions in interesting ways.
While weight-clipping might seem disadvantageous, directly programming a very ``long-tailed'' weight distribution by mapping its largest outlying weight-value to $\gmax$ can cause even larger problems.  
Such mappings tend to produce low average output currents which fail to employ the available ADC range, leading to larger \ac{MVM} errors thanks to \ac{ADC} quantization, output noise, and other nonidealities that remain stubbornly independent of the reduced output signal-levels. 

\begin{figure}[t]
  \centering
  \includegraphics[width=\textwidth, clip, trim=1.5cm 0.1cm 0 0cm]{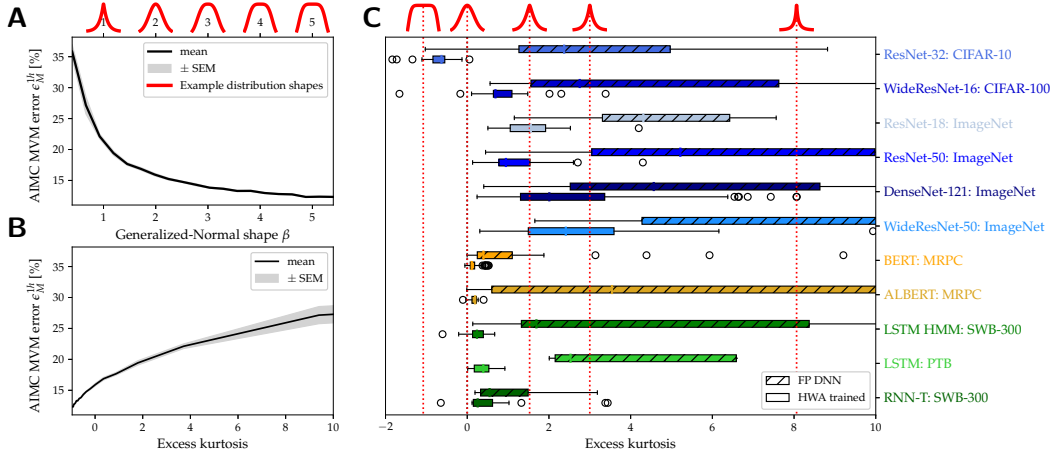}
  \caption{\small \ac{HWA} training reduces \ac{MVM} error by creating more compact conductance distributions. 
    (A) \ac{MVM} error decreases as constructed conductance distributions, produced by a generalized normal distribution (\eqref{gennorm}), are made more compact by increasing $\beta$.
    Example distributions in red at top show  $\beta=1$ (Laplace distribution), $\beta=2$ (normal distribution), and even more compact distributions for higher $\beta$. `SEM` indicates standard-error of the mean.
    (B) Data from (A) is replotted as a function of the (excess) kurtosis of the distribution. According to the definition of excess kurtosis, a normal distribution (that is $\beta=2$ in (A)) has a value of 0, and positive or negative values for longer tail distributions (ie. $\beta < 2$) or more compact distributions (ie. $\beta > 2$), respectively. Note that longer tail distributions (large kurtosis) lead to higher \ac{MVM} error, while more compact distributions (lower kurtosis) reduce \ac{MVM} error    
    (C) Kurtosis of the conductance values per layer, comparing \ac{HWA} trained models (solid bars),  to \fp\ weight data scaled by the overall absolute maximum weight (hashed bars).  Column-wise scaling, and the tuning of both weights and
    scaling-parameters during \ac{HWA} training, help lead to significantly more compact distributions with smaller kurtosis values. 
    }
  \label{fig:kurtosis}
\end{figure}

To show this effect, we calculate the \ac{MVM} error for different arbitrarily-constructed weight distribution shapes, obtained by sampling the generalized normal distribution,
\begin{equation}
  \label{eq:gennorm}
  p(x | \mu, \alpha, \beta) = {\frac {\beta }{2\alpha \Gamma (1/\beta )}}\;e^{-(|x-\mu |/\alpha )^{\beta }},
\end{equation}
where we use $\alpha=1$ and $\mu=0$. 
As $\beta$ increases, this distribution becomes more compact, moving through the Laplace ($\beta=1$) and normal distributions ($\beta=2$) along the way (see red curves above \figref{kurtosis}A). 
\figref{kurtosis}A shows the \ac{MVM} error $\mvmerror$ at 1 hour drift, for weight-values sampled from \eqref{gennorm} as $\beta$ increases from long-tailed ($\beta\le$1) to compact (high $\beta$) weight distributions.  Here we map weights directly to conductance values, with the maximum weight assigned to $\gmax$; inputs are uniformly distributed between $(-1, 1)$. 
\ac{MVM} error increases rapidly for longer-tailed distributions ($\beta\le1$).

One simple measure of a distribution's shape is the \emph{kurtosis}, obtained by dividing the fourth moment ($\langle (x - \mu)^4 \rangle$) of the distribution by its variance squared ($[ \langle (x - \mu)^2 \rangle ]^2$). 
In the plots and the remainder of this section we use the \emph{excess} kurtosis --- defined as the kurtosis minus 3, so that its value is 0 for normal distributions. Since kurtosis increases for long-tailed distributions, we find that lower kurtosis -- and thus more compact weight distributions -- means lower \ac{MVM} error (\figref{kurtosis}B). 

Fortunately, our \ac{HWA} training and conductance mapping approach tends to inherently produce more compact conductance distributions, for several different reasons. 
First, the individual digital scales $\affinescale_i$ available for each \ac{MVM} output (see \eqref{periphery}) are initialized to scale conductances by the absolute maximal value of each weight matrix-column rather than by the overall maximum across the entire weight matrix. 
With each column individually scaled, the overall conductance distribution becomes more
compact than the original weight distribution.  
During \ac{HWA} training, these digital scales are optimized -- which may lead the system to choose to clip some output columns -- and any large weight deviations and outliers created during training are also clipped. 
Finally, since the \ac{AIMC} nonidealities cause large weights and outputs to increase the errors that \ac{SGD} is attempting to correct, \ac{HWA} training should be expected to drive towards more compact weight distributions during retraining.

Indeed, we find that our \ac{HWA} training and mapping scheme greatly increases the compactness of the conductance distributions for each layer, as indicated by the kurtosis values shown for our 11 \ac{DNN} models in \figref{kurtosis}C. 
Hashed bars show kurtosis for direct mapping of the \fp\ model without \ac{HWA} training, using a single global digital scale-factor per layer. 
Solid bars illustrate that our columnwise-scaled and \ac{HWA}-trained models get mapped into conductance distributions that are \emph{significantly more compact}, which helps reduce both \ac{MVM} and \ac{DNN} error.

\subsection{Improving AIMC fidelity of selected  layers to reach iso-accuracy in large CNNs }

\begin{figure}[t]
  \centering
  \includegraphics[width=\textwidth, clip, trim=1.5cm 0.1cm 0 0cm]{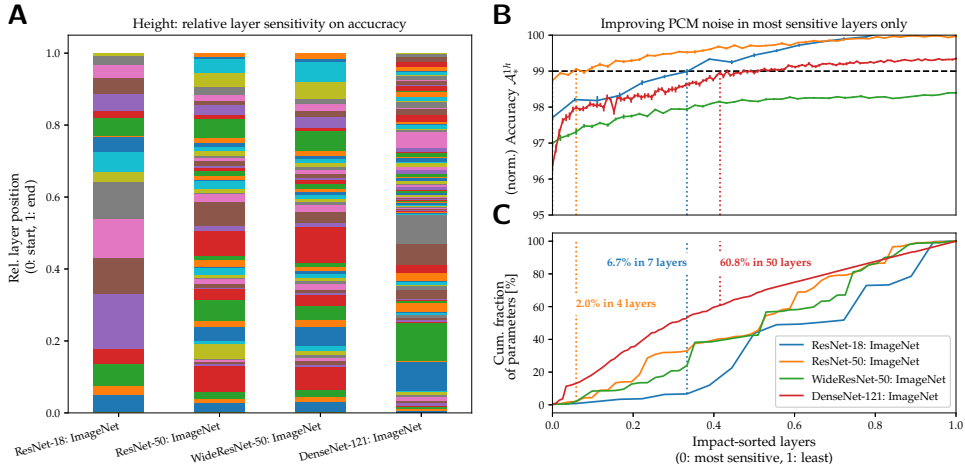}
  \caption{\small Layer-wise breakdown for ImageNet \acp{CNN}. 
    (A) Bar-charts reveal the relative impact that different \ac{DNN} layers have on \ac{AIMC} accuracy when the \ac{PCM} conductances in just that layer are made very noisy (overall \ac{PCM} noise scale set to $15$), while all other layers see only minimal \ac{PCM} noise (overall noise scale set to $0$).  
    The height of each bar-segment, arranged in sequential \ac{DNN} layer order, corresponds to the relative impact of that layer; colors simply delineate layer boundaries. 
    Note that ResNet-50 and WideResNet-50 have very similar graphs since their layers only differ in width. 
    (B) Accuracy $\accfphour$ as the most critical layers in these four \acp{CNN} are exempted from \ac{PCM} noise, plotted as the fraction of noise-exempt layers is increased, in order from most-sensitive to least-sensitive.  (C) Corresponding cumulative fraction of weight-parameters that are \ac{PCM}-noise-exempted.
    For ResNet-50 and ResNet-18, reducing \ac{PCM} noise in just a few layers (dotted vertical lines) allows an \ac{AIMC} crossbar-model to achieve iso-accuracy (dashed horizontal line).  
    For DenseNet-121, more than half of the network parameters need to be exempted, while for WideResNet, even the non-\ac{PCM} nonidealities would need to be improved.
    }
  \label{fig:imnet-ranked}
\end{figure}

Our results show that larger \ac{CNN}s, particularly those using the ImageNet database, are the most challenging for \ac{AIMC}.
Even with \ac{HWA} training, our standard \ac{AIMC} crossbar-model cannot achieve iso-accuracy for these \ac{DNN} models (\tabref{hwa-results}).
Clearly the fidelity of the \acp{MVM} must be further improved, either through better materials or through hardware design-choices.
For instance, designers could dedicate multiple conductance pairs per weight~\cite{le2022precision} to reduce \ac{PCM} programming errors, but at the cost of larger tile area and energy.  
Or designers could average the results from multiple passes through the tile to reduce the effects of cycle-to-cycle \ac{PCM} read and additive output noise, but at significant cost to latency, throughput, and energy-efficiency.  
Given these unpleasant tradeoffs, such approaches should be used as infrequently as possible, ideally only on a small set of \ac{DNN} layers that really require these extra resources, which can then allow the entire model to achieve iso-accuracy.

Thus we need to determine which of the layers in ImageNet \acp{CNN} are the most sensitive to \ac{AIMC} nonidealities, and then assess whether improving just a small subset of these layers would have sufficient impact.
To do this, we sequentially introduce \ac{AIMC} nonidealities at each layer of the \ac{HWA}-trained \acp{DNN} individually, while turning off all nonidealities in all other layers (using \fp\ operations on their \ac{HWA}-trained weight matrices).  
By repeating this process over the $L$ layers with different overall \ac{PCM} noise settings, we can determine the sensitivity and relative importance of single layers.

We first rank the layers according to accuracy impact for each \ac{DNN} by exposing each layer to significant \ac{PCM} noise with all other layers exempted from noise (\figref{imnet-ranked}A).  
Then, in order from most- to least-sensitive layer, we introduce this noise-exemption into multiple layers (\figref{imnet-ranked}B), causing normalized accuracy at 1 hour drift $\accfphour$ with respect to the \fp\ model to increase as more and more model-parameters are made noise-exempt (\figref{imnet-ranked}C). Eventually the 99\% iso-accuracy is achieved (dashed horizontal line) and then exceeded for most of these models.
For \figref{imnet-ranked}A, the one layer being assessed sees 15$\times$ the usual \ac{PCM} noise; for \figref{imnet-ranked}B, the layers not yet \ac{PCM}-noise-exempted see our standard \ac{AIMC} crossbar-model.  
While \ac{PCM}-noise-exempt layers experience no long-term conductance noise, programming errors, or drift, they still are subject to the same cycle-to-cycle read noise, additive output noise, and \ac{DAC}/\ac{ADC} quantization choices in our standard \ac{AIMC} crossbar-model. 

For ResNet-18 and ResNet-50, we find that improving just a few layers can help achieve iso-accuracy ($\accfphour \ge 99$\%, dashed line in \figref{imnet-ranked}B). 
This involves only $7$\% and $2$\% of the model parameters, respectively (\figref{imnet-ranked}C). 
Improving \ac{MVM} fidelity for such a limited number of parameters should prove less costly than across the full \ac{DNN}.
However, for the other two ImageNet \acp{DNN}, a considerably larger fraction of the model ($> 60$\%) would need to be improved to reach iso-accuracy. 
Therefore, these \acp{DNN} will either need further advances in either \ac{HWA}-training or the overall \ac{AIMC} specifications, in order to support \ac{AIMC} deployment without significant accuracy drop.

\section{Discussion}
\label{sec:discussion}

We have introduced a new approach for successfully deploying \acp{DNN} onto realistic \ac{AIMC} inference hardware, at or near iso-accuracy.
Our standard \ac{AIMC} crossbar-model incorporates well-known but hardware-calibrated nonidealities caused by the analog devices, such as read noise, programming errors, and conductance drift.   
Going well beyond previous studies, our model also includes nonidealities due to \ac{MVM} circuit integration, such as
system noise, \ac{DAC} and \ac{ADC} quantization, and dynamically-computed IR-drop. 
Finally, our model fully addresses the fixed dynamic-range constraints on inputs, weights, and outputs found in all \ac{AIMC} systems, but previously neglected.

While a few aspects of this study are not directly applicable to hardware designed around non-\ac{PCM} devices,
our standard \ac{AIMC} crossbar-model and our carefully-designed inference protocols can readily serve as the basic core for studying such systems.  
The intuition we have developed in terms of how various types of noise affect different families of \ac{DNN} models is also readily transferable.
As such, the present work establishes a baseline that can both guide -- and be compared against -- future \ac{AIMC} simulation studies.
To help make this even more straightforward, our standard \ac{AIMC} crossbar-model has now been incorporated into our open source \ac{AIHWKIT}~\cite{Rasch2021AFA}, which is based on the popular ML framework PyTorch~\cite{pytorch}.

Our \ac{AIMC} crossbar-model is not a replacement for the detailed circuit simulations essential to hardware verification.
We use many simplifications and abstractions of the various \ac{AIMC} nonidealities, since our goal is quick and relatively realistic functional verification of larger \ac{DNN} workloads. 
For instance, we assume noise sources are Gaussian, avoiding physically-modeled distributions that would be more accurate but significantly slower. We also devised a method to rapidly approximate IR-drop which can adjust dynamically with the
input. 
We chose to intentionally ignore static crossbar effects that would change the conductance value systematically~\cite{shubham2021, roy2021}, since read-write-verify conductance programming can readily adapt to such effects.

Some prior works proposed using on-chip or chip-in-the-loop training methods~\cite{Joshi2020, wan2022compute, shubham2021, yao2020fully, wright2022deep}, which can greatly increase the attainable accuracy by addressing the specific fabrication variations found on that particular chip. 
However, we strongly believe that the time and cost of such individualized preparation is likely to be untenable for widespread deployment.   
Thus in this paper, we have focused on \ac{HWA} training that can be general enough to be performed once per model per \ac{AIMC} chip-family, greatly simplifying the deployment onto individual chips. 
That said, our \ac{HWA} training approach could readily be combined with more sophisticated online compensation methods, with on-chip or chip-in-the-loop training, or with more than one device pair used per weight, including optimization of how weights are assigned across these conductances~\cite{mackin2022optimised}.

Since \ac{HWA} training is performed in software before deployment, it has no first-order impact on the latency, throughput or energy-efficiency of \ac{AIMC} hardware. 
However, as we have shown, \ac{HWA} training is essential to \emph{understanding the tradeoffs} between accuracy and these important system performance metrics. 
 For instance, because of the sequential nature of layers of a deep network, shallower but wider layers should
generally be preferable for \ac{AIMC}, since higher utilization of large matrices stored on the crossbar arrays does not significantly change the runtime~\cite{gokmen2016acceleration, rasch2019rapa} and helps improve energy-efficiency. 
In terms of noise robustness, excessively deep \acp{DNN} have disadvantages. 
Among the ImageNet \acp{CNN} tested, DenseNet-121 showed the worst accuracy drop from its \fp\ model,
while WideResNet-50 offered the best raw test-error at 1 hour drift (23.82\%, versus 24.82\% for ResNet-50, \tabref{hwa-results}). 
Together with information on the latency, throughput, and energy-efficiency, this kind of information on available accuracy gain
is critical when trying to decide which \ac{DNN} model to deploy.

A few previous studies have attempted to improve the robustness of \acp{DNN} to nonidealities by noise-aware training, where multiplicative or additive Gaussian noise~\cite{Kariyappa2021,Joshi2020}) is added to weights or activations during training.
Similarly, other studies seeking to prevent overfitting or to enhance
robustness to adversarial attacks have injected noise into standard floating-point
training as a regularization technique~(\cite{dropout2014, wager2013, goodfellow2013, kang2016, noh2017, rakin2018, Li2020}). While all these methods qualitatively increase the noise robustness of \acp{DNN}, the quantitative benefits on real \ac{AIMC} can neither be accurately reported nor fully optimized by these studies. Since our \ac{HWA} approach keeps weights mapped in conductance units, a variety of realistic hardware-relevant constraints can be incorporated in a straightforward manner.
These include the complexities of \ac{PCM} programming, and the shallow input-output ranges, IR-drop and quantization affecting the \ac{MVM} compute -- aspects neglected in most previous studies.

We have tried distilling with the \ac{FP} model as a teacher (similar to~\cite{zhou2020noisy}) and found some benefits when \ac{HWA} training time is limited. However, since the improvements offered by distilling disappeared at longer training times for most \ac{DNN} models, we mostly report results without distilling.  We did find that accuracy with distilling is significantly higher for the \ac{HMM} Speech \acs{LSTM}, and these results are shown in \tabref{hwa-results},  implying that distilling can be helpful for some \acp{DNN}.

Rather than simple Gaussian weight noise~\cite{Joshi2020}, we use the expected weight noise distribution characterized from \ac{PCM} measurements~\cite{Nandakumar2019}.   
We find that applying readout-noise on the weights -- together with the correct incorporation of injected longer-term programming-noise when modifying the weight matrix during the backward pass -- is crucial for achieving \ac{AIMC} robustness. 
One drawback of our approach is that this type of noise injection is currently applied only once per mini-batch, which reduces the effectivity of the noise as batch-size increases.  
One possible improvement would be to sample the weight noise sources multiple times per mini-batch. Such an extension of our methods should further improve the noise robustness of the \ac{HWA}-trained \acp{DNN}.

\section{Conclusion}

In summary, we show that comprehensive \acf{HWA} training can greatly enhance the robustness of a variety of \acfp{DNN} -- including \acfp{CNN}, \acfp{RNN}, and transformers -- to the unavoidable device and circuit nonidealities of emerging \acf{AIMC} accelerators.
In five of the 11 models studied, the techniques we introduce lead to software-equivalent accuracy, defined as 99\% of the accuracy-performance offered by the original \ac{DNN} model beyond random guessing.  
Across all models, \ac{HWA}-training reduces the worst-case gap in raw model accuracy from 21.81\% down to just 2.65\%.

Through a systematic sensitivity analysis, we identify the nonidealities that are most critical for maintaining accuracy in future system designs. 
For instance, we observe that nonidealities that effectively \emph{add noise to the inputs or outputs} -- such as \ac{ADC} and \ac{DAC} resolution, additive output noise, and S-shaped non-linearity of the ADC -- have the largest impact on \ac{DNN} accuracy. 
We also show that certain \ac{DNN} topologies, such as \acp{RNN} or shallower \acp{CNN}, can tolerate more \ac{AIMC} nonidealities than others. 

By making this standard \ac{AIMC} crossbar-model available in the open-source \ac{AIHWKIT}~\cite{Rasch2021AFA}, we make it possible for future advances in \ac{HWA} training techniques to be readily compared to these results.
By pinpointing the measures needed to compensate for imperfect \ac{AIMC} hardware, the tools we have introduced here enable better understanding and optimization of the tradeoffs between model accuracy and desirable performance characteristics such as latency, throughput, and energy-efficiency.  
Continued coordination between \ac{HWA}-training and architectural assessments may even lead to brand-new \ac{DNN} topologies,  specifically designed to maximize the benefits of \ac{AIMC} hardware --- accurate inference at high speed and low power.

\section*{Methods}

\subsection*{AIMC standardized evaluation model}

\subsubsection*{Affine transform in tile periphery}

We assume that each output column of the analog crossbar has a
floating-point scale $\alpha_i$ and offset $\beta_i$ available which implement
together an \emph{affine transformation}. We assume that conductances can be linearly
mapped to weight values, so that we can normalize the analog weight values from
-1 to 1, corresponding to  $-\gmax, \ldots, \gmax$ (see conductance
programming below). This affine transform then maps the column's physical output
(e.g. current), as quantized using an \ac{ADC} into integers within a
certain range, to the value expected by the DNN for the next layer
(e.g. activation). Note that such \ac{ADC} conversion using a scale and bias per
column is already available in prototypes~\cite{khaddam2021hermes} but has not previously been incorporated into studies on
\ac{HWA} training.

This digital periphery of an analog \ac{MVM} can thus summarized at
\begin{equation}
  \label{eq:periphery}
  \yaimc_i = \affinebeta_i + \sinput \affinealpha_i\, \quant_{\bout}^q
  \left(\fanalogMVM\left(\quant_1^q\left(\mathbf{x}/\sinput\right)\right)\right),
\end{equation}
where $\fanalogMVM$ describes the analog aspects of the \ac{AIMC} \ac{MVM} (see ``Analog MVM model''), and
\begin{equation}
  \label{eq:quant}
  \quant_b^q(z) \equiv \clip{-b}{b}{\round\left(\frac{(2^q - 2) z}{ 2 b}\right)},
\end{equation}
describes linear quantization to $2^q-1$ values in $-b, \ldots, b$ centered around 0.
One bin is discarded to force an odd number of bins on either side of zero.  
Here  $\clip{a}{b}{x}$ constrains $z$ between minimum $a$ and maximum $b$,
\begin{equation}
  \label{eq:clip}
  \clip{a}{b}{z} = \begin{cases}
                     z, & \text{if } a < z < b \\
                     a,  & \text{if } z \le a \\
                     b,  & \text{if } z \ge b.
                   \end{cases}
\end{equation}
$\sinput$ is a scalar, per-crossbar value which determines the usable input range.
This can either be a learned parameter which is then held fixed during inference (static input range), 
or can depend dynamically on the current input vector $\mathbf{x}$ (dynamic input range). 
While main results assume a static input range, we examine performance improvements for the dynamic option (Supplemental Material, \secref{input-range-learning}).

The scales $\affinealpha_i$ determine the mapping of conductances to weight values, individually for each crossbar column $i$. 
During \ac{HWA} we allow \ac{SGD} to optimize this parameter, starting from values initialized as described below. $\affinebeta_i$ is used to implement the bias of the \ac{MVM}, which we implement in digital (\ac{FP}) precision here.
We assume $q=8\bit$ quantization, and investigate lower precision as part of our sensitivity analysis.

\subsubsection*{Dynamic MVM range}

A critical feature of our crossbar-model is that it fully encompasses the finite dynamic-range constraints on inputs, weights and outputs that will be present and unavoidable in any real \ac{AIMC} implementation.
Since both input and weights are normalized within $-1,\ldots 1$ (in analog units), our output-bound setting of $\bout=10$ means that just 10 fully-on inputs, applied to rows containing maximal-value weights, would fully saturate the output. 
This is a conservative choice that works for modest-size crossbars and for our assumption that positive current contributions (produced by weight and activation pairs of the same sign) and negative contributions (weights and activations have opposite signs) cancel \emph{within} the array.
This mode is energy-efficient and minimizes IR-drops, but requires the \ac{ADC} to be capable of measuring bipolar currents~\cite{khaddam2021hermes}.  
If the crossbar is made much larger, or the positive and negative terms are integrated separately, this may increase energy usage and exacerbate IR-drops, but simplify the \ac{ADC} design. Furthermore, such choices will likely alter the overall dynamic range limitations, calling for a reoptimization of $\bout$.

\subsubsection*{Analog MVM model}
Our basic model is illustrated in \figref{aimc-mvm-model}A. The analog \ac{MVM}
$\yyana = \fanalogMVM(\xxana)$ in \eqref{periphery} for the quantized, clipped
and scaled input vector $\xxana \equiv \quant_1^q(\mathbf{x}/\sinput)$ takes the
following general form
\begin{equation}
  \label{eq:analog-mvm}
  \yana_i =  \sigout\xi_i + \fnonlin_i\left(\yirdrop_i + \sum_{j}\left(\wana_{ij}(\tinf)+
    \sigwnoise(\wana_{ij})\, \xi_{ij} \right) \,  \xana_j \right),
\end{equation}
where analog weights $\wana_{ij}(t)$ represent normalized conductances with programming errors, drift, and long-term noise up to time $\tinf$ applied (see `Weight programming' below). 
We include a point-wise non-linear functions $\fnonlin_i(x)$ to support special cases such as \ac{ADC} nonlinearities; in our standard model, $\fnonlin_i(x)\equiv x$. 
Normal random numbers ($\xi_i, \xi_{ij} \sim {\cal N}(0, 1)$ are drawn for each \ac{MVM}, representing additive output noise with standard deviation $\sigout=0.04$, and short-term weight noise $\sigwnoise(\wana)$ that depends on the current weight values (see `Short-term \ac{PCM} read noise'), respectively.  
Since the analog output values running from $-10,\ldots 10$ get quantized into digital values from $-127,\ldots 127$ ($8\bit$), this choice of $\sigout=0.04$ corresponds to almost exactly half of one \ac{ADC} quantization bin.

\subsubsection*{Weight programming}

We adopt a previously-described and -characterized weight-programming and drift model for \ac{PCM} devices~\cite{Nandakumar2019}. 
We assume that the crossbar provides 1 pair of conductances per weight, where the first (second) member of the device-pair is programmed to a conductance between reset (0) and set ($\gmax$) to handle positive (negative) weights, with the non-active conductance programmed to reset.  Only the active conductance is considered in our model.
Although recent prototypes support two pairs per weight~\cite{narayanan2021vlsi, khaddam2021hermes}, having only one conductance-pair increases the weight density and thus compute efficiency, and poses a more difficult challenge in terms of accuracy and yield. 

Each column $\mathbf{w}_i$ of each weight matrix is mapped to a column of target conductances $\ggtarget_i$.  We first initialize each affine scale coefficient using the maximum weight found in that column, $\affinealpha_i = \max_j |w_{ij}|$.  This allows each weight to be mapped to a scaled target conductance, $\gtarget_{ij} = \gmax\frac{w_{ij}}{\affinealpha_i}$. In our \ac{HWA} training approach (described below), after this initialization of target conductance and affine scales based on the \fp\ model weights, we then use \ac{SGD} to further optimize \emph{both} the mapped target conductances and scales $\affinealpha_i$ separately. \tabref{hwa-results} uses this learned weight-to-conductance mapping when evaluating \ac{AIMC} inference performance.   

In a real \ac{AIMC} system, a positive $\gtarget$ value gets programmed onto a different physical device than if that particular $\gtarget$ had been negative. We here assume that only one of the two devices are programmed to particular target conductance whereas the other device is always at reset conductance ($\gtarget_{ij}=0$). In this case, one can simplify and compute the \ac{MVM} directly with signed conductances as done in our model. 
The programmed conductances $\gprog_{ij}$ differ from the desired target values $\gtarget_{ij}$ as
$\gprog_{ij} = \gtarget_{ij} + \sigprog(\gtarget_{ij})\,\xi$
due to programming noise, assumed to be Gaussian ($\xi \in \mathcal{N}(0, 1)$). 
In turn, the standard deviation of this programming noise depends on the target conductance as 
\begin{equation}
  \label{eq:prog_noise}
  \sigprog(\gtarget) = c_0 + c_1 \frac{\gtarget}{\gmax} + c_2 \frac{\gtarget^2}{\gmax^2},
\end{equation}
where $c_0=0.26348\,\mu\text{S}$, $c_1=1.9650\,\mu\text{S}$, and $c_2=-1.1731\,\mu\text{S}$, as obtained by fitting to extensive PCM hardware data~\cite{Nandakumar2019}.


\subsubsection*{Weight drift and read noise}

Once a \ac{PCM} device is programmed, the device exhibits both conductance
drift and $1/f$ (long-term) read noise. 
As briefly described below, both are modeled in a statistical manner based on measurements of doped-Ge$_2$Sb$_2$Te$_5$ (d-GST) mushroom \acp{PCM} from a large device array integrated in 90nm CMOS technology~\cite{Nandakumar2019}.

\paragraph{\ac{PCM} drift}
\ac{PCM} conductance drift, attributed to post-programming structural relaxation, follows an empirical relation
\begin{equation}
  \gdrift(\tinf) = \gprog\left(\frac{\tinf + t_0}{t_0}\right)^{-\nu},
  \label{eq:pcm_drift}
\end{equation}
where $\gdrift(\tinf)$ is the conductance measured at time $\tinf$
after the programming (assumed to complete at $t_0 = 20$s~\cite{Nandakumar2020_DV}) and $\nu$ is the drift coefficient. 

The drift coefficients for each device are assumed to be normally
distributed, that is
$\nu_{ij} \in \mathcal{N}\left(\mu_\nu(\gtarget_{ij}),
  \sigma_\nu(\gtarget_{ij})\right)$, where the mean and standard
deviation are empirically determined by fitting to experimental
data. The data fits are expressed by a clipped linear function in
log-space, that is (with \eqref{clip})
\begin{equation}
  \label{eq:clipped_lin}
  L\left(x | a, b, \ymin, \ymax \right) \equiv \clip{\ymin}{\ymax}{ a \ln x + b}
\end{equation}
where here $x\equiv \frac{\gtarget}{\gmax}$. The parameters for $\mu_\nu$ are
given by $a=-0.0155$, $b=0.0244$, $\ymin=0.049$, and $\ymax=0.1$. For
$\sigma_\nu$ the parameter are $a=-0.0125$, $b=-0.0059$, $\ymin=0.008$, and
$\ymax=0.045$.  The drift coefficient $\nu_{ij}$ thus determined for each device are
used to model the conductance at any time $\tinf$ using \eqref{pcm_drift}.

\paragraph{\ac{PCM} read noise}
\ac{PCM} is also known to demonstrate low frequency noise such as random
telegraph noise (RTN) and $ 1/f^\gamma $ noise with $ \gamma \in [0.9, 1.1]$. We
follow the empirical noise model of~\cite{Nandakumar2019}, which assumes
$\gamma=1$ and arrives at a read noise standard deviation at time
$\tinf$ of (\cite{Nandakumar2019})
\begin{equation}
  \label{eq:readnoise}
  \sigma_{\text{read}}(\tinf) = \gtarget\, Q_s(\gtarget) \sqrt{\ln\left(\frac{\tinf+T_{\text{read}}}{2\,T_{\text{read}}}\right)},
\end{equation}
where $Q_s(\gtarget)$ is measured to be
\begin{equation}
  \label{eq:qs}
  Q_s(\gtarget) = \clip{0}{c_3}{c_1\left(\frac{\gtarget}{\gmax}\right)^{c_2}},
\end{equation}
with $c_1=0.0088$,  $c_2=-0.65$, $c_3=0.2$.

This read noise is added to the post-drift conductance $\gdrift(\tinf)$ to arrive at the final \ac{PCM} conductance

\begin{equation}
  \label{eq:final}
  \gfinal = \clip{\gmin}{\infty}{\gdrift(\tinf) + \sigma_{\text{read}}(\tinf) \xi}
\end{equation}
where we set $\gmin=0$ here and $\xi \sim {\cal N}(0, 1)$. The weight
values $\analog{w}_{ij}$ of the crossbar array for \eqref{analog-mvm}
are then obtained by scaling and combining positive and negative parts
\begin{equation}
  \label{eq:final-weights}
  \analog{w}_{ij} = \frac{\gfinal_{ij}}{\gmax} \sign w_{ij}
\end{equation}
These long-term \ac{PCM} effects are applied to all weights prior to the evaluation at time $\tinf$ and the weights are
subsequently fixed during the evaluation of the test set. 
Short-term weight noise, redrawn for each \ac{MVM}, is included separately
in \eqref{analog-mvm} as described in the following paragraph.

\paragraph{Short-term \ac{PCM} read noise}
When evaluating the \ac{AIMC} \ac{DNN} at a time $\tinf$, the analog
weights $\analog{W}$ are established as described in
\eqref{final-weights}. However, weights are often re-used multiple
times during a single input, say across image-pixels in a \ac{CNN} image or sequence-tokens in
an \acp{RNN} or transformer model. 
Here short-term weight noise can cause small but perceptible cycle-to-cycle variations (\figref{aimc-mvm-model}B).

Modifying the weight matrix at each \ac{MVM} would be highly inefficient for our \ac{HWA} training
software running on GPUs.
To efficiently model such short-term read noise, we use the read noise definition \eqref{readnoise} to set $\sigwnoise$ in \eqref{analog-mvm}, but refer the resulting noise to the output $\yana_i$. 
Assuming zero-mean independent normal distributions, we can sum the variances as
\begin{equation}
  \label{eq:short-term-wnoise}
  \sigwnoiset_i = \sigwnoise_0\sqrt{\sum_{j}|\wana_{ij}|\,|\xana_j|^2},
\end{equation}
implying that the weight dependence of the read noise can be approximated as $\propto \sqrt{|\analog{w}|}$. 
Thus weight-noise $\sigwnoise$ in \eqref{analog-mvm} effectively adds $\xi_i\sigwnoiset_i$ (with $\xi_i\sim {\cal N}(0, 1)$) to the analog output $\yana_i$.  
The parameter $\sigwnoise_0$ can be identified with $c_1 \sqrt{\ln(\frac{\Delta t + t_{\text{r}}}{2t_{\text{r}}})}$ for read noise accumulated over time-period $\Delta t$ (\eqref{readnoise}, \cite{Nandakumar2019}). 
Assuming a read duration of $t_\text{r} = 250$ns and approximate waiting time between two consecutive \acp{MVM} ($\Delta t$) to be $100\times$ longer, we find $\sigwnoise_0 \approx 0.0175$.

\subsubsection*{Drift compensation}
For evaluation times $\tinf$ long after \ac{NVM} programming, the conductance drift \eqref{pcm_drift} can be compensated in the digital domain without any expensive re-programming~\cite{LeGallo2018_CS,Ambrogio2019}. 
This can be done by running a number of analog \acp{MVM} on some known test inputs $\{\mathbf{x}^{k}\}$ immediately after weight programming and recording the overall output magnitude as $s_\text{ref} = \sum_{ik}|y_i^{(k)}|$.  
At time $\tinf$, just before beginning inference, the same inputs can be applied to measure $s_{\tinf}$. 
We then correct the \ac{MVM} outputs by adjusting the digital $\affinescale_i$ (see \eqref{periphery}) by  $\frac{s_\text{ref}}{s_{\tinf}}$ to accommodate the average conductance decrease due to drift. 
We assume one global drift compensation applied to all columns, 
although this could be done individually at each column if $s_\text{ref}|_i$ can be measured sufficiently accurately. 
Other more sophisticated drift compensation and adaptive refresh methods including in-memory re-training could potentially be applied as well~(e.g.~\cite{Joshi2020}). 

\subsubsection*{Crossbar tile size}
The \ac{NVM} crossbars available on an \ac{AIMC} chip are of finite size, typically ranging from $256\times256$ (e.g.~\cite{khaddam2021hermes}) to $512\times 512$ (e.g.~\cite{narayanan2021vlsi}). 
We assume a tile-size of $512 \times 512$, and assume that enough crossbars are available to
support separate crossbars for each weight matrix.  
Any weight matrix with input dimension $>512$ is divided into roughly equal parts for programming on as many tiles necessary.  Partially-used tiles have weights are situated at the bottom of the crossbar, to minimize interference and potential IR drop, and unused inputs are clamped to zero.

Each tile computes an \ac{MVM} (\eqref{analog-mvm}) using its own periphery (\eqref{periphery}). 
Inter-tile summation is performed at \ac{FP} precision (FP16), after affine-scaling but before being passed to subsequent digital compute such as activation functions.  
Because our \ac{AIMC} nonidealities have no dependencies across output columns, the \ac{HWA} training code does not need to explicitly break the compute along the output-dimension into tile-sized chunks. This helps the simulations run more efficiently on GPUs.

\subsubsection*{IR-drop}
Ideally, the voltage along each long bitline in the crossbar would remain constant, so that conductances with the same value could contribute the same current, whether in the farthest or nearest row from where peripheral circuitry is holding the bitline voltage and measuring currents. 
In a physical crossbar, however, IR-drops imposed by finite wire resistance cause the bitline voltage to vary~\cite{chen2013comprehensive}, especially as instantaneous currents get large.
To keep the simulation-time reasonable, we make a number of approximations when modeling this effect. 
IR-drop is modeled independently for each crossbar-column, because any column-to-column differences will be implicitly corrected (to first order) when programming the weight with an appropriate read-write-verify scheme.  

However, within each crossbar column, the current contributed by each weight depends on the local bitline voltage, which in turn depends on the other currents being generated elsewhere along the column by that particular input vector.
This situation will evolve throughout the integration period due to the pulse-length modulation of those inputs as well as any resulting transients, including the response of the peripheral circuit establishing the bitline voltage.
Here, for simplicity and speed of computation for large \acp{DNN}, we only consider the average integration current. 

The steady-state bitline voltages $\bar{v}_i$ can be computed by solving the equation system
\begin{equation}
  \label{eq:ir_system}
  \left(\bar{v}_{i+1} - \bar{v}_i\right)\,g_w + g_i^+(v_i^+ - \bar{v}_i) = \left(\bar{v}_i- \bar{v}_{i-1}\right)\,g_w+ g_i^-\left(\bar{v}_i - v_i^-\right)
\end{equation}
where $g_w$ is the wire conductance between the crosspoint nodes and $g_i^{+/-}$ the weight programmed onto either the positive or negative conductance (with the other programmed into the reset condition, $g=0$).
The individual input voltages, $v_i^-$ and $v_i^{+}$ of spatially-ordered inputs $i$, are linearly prorated from the supply voltages ($v_\text{ref} \pm V_\text{read}$) to represent the time-averaged current. The analog output current $\yana$ located at location $i=0$ is given by $g_w\left(\bar{v}_0 - v_\text{ref}\right)$, with $V_\text{read} = 0.2$V. 
  
This linear system (\eqref{ir_system}) can be solved by inverting the unique coefficient matrix produced by a given input vector. 
To speed up the simulation and avoid inverting a $512\times512$ matrix for each \ac{MVM}, we further approximate the solution with a quadratic equation. 
Thus, in our analog \ac{MVM} (\eqref{analog-mvm}), the IR-drop amount is computed from the normalized weights and inputs by
\begin{eqnarray}
  \label{eq:irdrop}
    a_i &\equiv& \gamma n \sum_j |\wana_{ij}| |\xana_j|\\
    c_i &\equiv& 0.05\, a_i^3 - 0.2 a_i^2 + 0.5 a_i \\
    \yirdrop_i &\equiv& - c_i\sum_j \wana_{ij} \xana_j \left(1 - (1 -
                        \frac{j}{n})^2\right), \label{eq:irdrop-final}
\end{eqnarray}
where $\gamma$ is the unitless product of the wire resistance between adjacent cross-points (assumed 0.35~$\Omega$) and the maximal (set) conductance of the device ($g_\text{max}=5\mu$S), and $n$ is the number of cross-points occupied by the weight matrix.
We assume that smaller weight matrices are located at the lower edge of the crossbar to avoid excess IR-drop. 
We use \eqref{irdrop-final} to dynamically approximate the IR-drop across the $512$ input channels in \eqref{analog-mvm} when computing normalized \ac{MVM} outputs $\yaimc$ in all our results. 
Multiplying these normalized outputs by $g_\text{max}V_\text{read}$ produces the (time-averaged) physical output currents.   
To amplify these IR-effects for the sensitivity analysis (\figref{sensitivity}), we simply multiply the IR-drop error $\yirdrop_i$ by a varying scaling factor.  

For large inputs where current is flowing throughout the integration window, our estimations using time-averaged current are quite accurate. 
However, for small inputs where much of the current-flow occurs in a small portion of the integration window, instantaneous and average currents differ strongly, and IR-drop will be underestimated.   
We find that for a Normal distributed weight matrix and random but correlated inputs (as in \figref{aimc-mvm-model}E), IR-drop deviations are underestimated by roughly a factor of 5. 
Unfortunately, similar conditions arise across many of our DNNs.  
Fortunately, our sensitivity analysis (\figref{sensitivity}) finds that scaling our time-averaged IR-drop approximation by a factor of $>10\times$ does not significantly impact the accuracy of the \acp{DNN},
so we can still conclude that \acp{DNN} are reasonably robust to IR-drop, albeit by a modest rather than large safety margin.
Since IR-drop depends heavily on both on the hardware design (crossbar size, wire resistances, and absolute device conductances) and on the input and weight distributions,  
detailed circuit-based simulations using the intended workload(s) will remain a critical part of assessing new hardware designs.

\subsection*{Additional nonlinearities for sensitivity analysis}

\subsubsection*{PCM device yield}
Emerging memory devices such as \ac{PCM} exhibit imperfect yield, and some fraction of the devices in a given crossbar array will simply not switch properly~\cite{WKim2016_ALD,chen2017}. 
\ac{PCM} devices can end up stuck-at-set ($\gmax$), stuck-at-reset (conductance set to 0) and stuck-at-random (stuck somewhere between 0 and $\gmax$). 
In our sensitivity analysis (\figref{sensitivity}), we vary the fraction of failed devices and randomly-select their locations.

\subsubsection*{S-shaped ADC output non-linearity}
The output level might gradually saturate more gradually than the desired linear response due to non-linearity in the ADC~\cite{tsai2018power, khaddam2021hermes}. 
To estimate the impact of this for our sensitivity analysis (\figref{sensitivity}), we define $\fnonlin_i$ in \eqref{analog-mvm} with
\begin{equation}
  \label{eq:s-shaped-adc}
  \fnonlin_i(z) \equiv \left(1 + \frac{2}{d_\text{out}}\sum_{k=1}^{d_\text{out}}{|\zeta_k|}\right)^2 \frac{z}{1 + |\zeta_i z|},
\end{equation}
which models a S-shaped saturation with variable slope scaled to approximately cover the full output range. 
Each of the $d_\text{out}$ outputs has an independent \ac{ADC} and thus a slightly different (pre-determined) shape, $\zeta_i = \mu_\zeta\,(1 + \sigma_\zeta \xi)$ with $\xi \sim \mathcal{N}(0, 1)$ and $\mu_\zeta = \frac{1}{4}$. 
$\sigma_\zeta$ is only varied in the sensitivity analysis (``\ac{ADC} S-shaped nonlinearity''); for our standard \ac{AIMC} crossbar-model,
$\mu_\zeta$ and $\sigma_\zeta$ are both set to 0, causing $\fnonlin_i(z) = z$.

\subsubsection*{PCM  polarity}
Depending on the hardware and unit-cell design, positive and negative inputs might not create perfectly symmetric read-currents. 
The measured conductance of a \ac{PCM}-device can depend on whether read-current passes from top to bottom electrode, or vice-versa.  
This read-polarity dependence can cause weights to appear systematically altered for negative inputs as compared to positive inputs.  
Although the average effect can be corrected by adjusting read voltages, device-to-device or conductance-dependent variations can remain. 
To model this effect in our sensitivity analysis, we separate positive and negative inputs into two phases (setting a negative input to 0 in the positive phase and vice versa), and scale each weight in the negative phase by $(1 + a_{ij})$ where $a_{ij} \sim \mathcal{N}(0, \sigma_a)$. 
We then vary this new nonideality parameter $\sigma_a$ as ``weight asymmetry std.''

\subsection*{MVM error calculation}

To quantify the fidelity of the analog \ac{MVM}, we calculate the expected deviation of the analog \ac{MVM} as compared to the ideal \ac{MVM} as \ac{MVM}-error $\mvmerror$, defined by the relative normalized deviations (see e.g.~\cite{buchel2022gradient})
\begin{equation}
  \label{eq:mvm-error}
  \mvmerror(W, \left\{\mathbf{x}_k\right\}) = \frac{\langle||\mathbf{y}_k - \yyaimc_k||_2\rangle_k}{\langle||\mathbf{y}_k||_2\rangle_k},
\end{equation}
where $\mathbf{y}_k=W\mathbf{x}_k$ is the ideal \ac{MVM} output to input vector $\mathbf{x}_k$ using matrix $W$, and $\yyaimc$ is the actual \ac{AIMC} output considering all hardware-related nonidealities as defined in \eqref{periphery}.

The \ac{MVM}-error is obviously zero if the \ac{AIMC} is equal to the ideal outcome, but otherwise it depends on both the particular weight matrix $W$ and set of input vectors $\mathbf{x}_k$ used to estimate \eqref{mvm-error}.
To best reflect the impact of the nonidealities on the \ac{DNN},  inputs $\mathbf{x}_k$ should ideally be taken from the distribution of actual input activation vectors, and $W$ should be the target weight matrix, for the specific \ac{DNN} layer in question. 

However, to quantify the \ac{MVM}-error independent of the \ac{DNN} in question, we calculate the standard \ac{MVM}-error $\mvmerrorstd$ by using normal distributed weights, $w_{ij} \sim \mathcal{N}(0, 0.246)$ and uniform inputs $x_i \sim \mathcal{U}(-1, 1)$ with a tile size of $512 \times 512$. 
For our standard \ac{AIMC} crossbar-model as described in `AIMC standardized evaluation model', the standard \ac{MVM} error is $\mvmerrorstd=15\%$ (not considering drift).

\subsection*{AIMC hardware-aware DNN training}

Robustness to the nonidealities of \ac{AIMC} inference hardware can be improved by \acf{HWA} training --- a \ac{DNN} re-training method that applies expected nonidealities to the forward pass of the \ac{SGD}, with the backward pass performed using regular \ac{FP} precision.

Our \ac{HWA} training approach is to use the general form of the expected analog
\ac{MVM} nonidealities as described in \eqref{analog-mvm} together with the
injection of the expected programming errors (but without any conductance drift). Further, we use the \ac{HWA} training step to also
establish the digital peripheral parameters of \eqref{periphery}, in particular
the static input range $\sinput$ (see `Learning the input range') and the
weight-to-conductance mapping $\affinescale_i$
(see `Learning of weight-to-conductance conversion factors'). Additionally, we found that ramping up the injected
programming error strength (see `Re-training with weight noise injection'), fixed scales and
individual learning rates per tile (see `Learning of weight-to-conductance conversion factors'), weight clipping
(see `Weight mapping and clipping') and distilling (see `Distilling with floating-point teacher') improved
the robustness and achievable accuracy in the presence of \ac{AIMC} nonidealities.

In general, the \ac{HWA} training starts from an already \ac{FP}-trained \ac{DNN}, and hyper-parameters (learning rate, injected noise strength) are optimized.  
We verified the effectiveness of our new \ac{HWA} training approach on the very same \acp{DNN} used in a previous study~\cite{Joshi2020} and found, on average, a $>10$\% decrease in \ac{AIMC} test error for long $\tinf$ times.  
This directly indicates the improvement of our approach over previous methods (see \tabref{hwa-validation}). 

In the following paragraphs, our new \ac{HWA} training methods are presented in more detail.

\subsubsection*{Re-training with weight noise injection}
Injecting noise to improve robustness to nonidealities was suggested by a number
of studies~(e.g.~\cite{Joshi2020, gokmen2019iedm, Kariyappa2021}), and has been
one of the hallmarks of \ac{HWA} training for \ac{AIMC}.
In previous studies, noise has been injected in multiple ways, such as
output~\cite{gokmen2019iedm, Joshi2020}, input~\cite{Joshi2020}, or weight
noise~\cite{Joshi2020,Kariyappa2021}. Different types of weight noise distributions have been used, such as
additive (scaled by the current maximal weight~\cite{Joshi2020}) or
multiplicative~\cite{Kariyappa2021} Gaussian.

Methods for injecting weight noise have differed across previous studies. 
For instance, Joshi et al.~\cite{Joshi2020} added newly drawn Gaussian weight noise to the weight matrix reversibly for each image input (not mini-batch) only during the forward pass (and not during backward pass which was done with the actual weight matrix). 
However, it is more mathematically-correct to also apply these same weight perturbations during the backward pass (but not to the reference weights to which updates are applied), as is commonly done for weight regularization techniques such as drop-connect~\cite{wan2013}.  
Furthermore, although the exact noise injection method (input, output, or weight noise) does not seem to matter much~\cite{Joshi2020}, generic additive Gaussian noise does not conform with the expected \ac{AIMC} noise structure. For instance, \ac{PCM} programming errors are actually conductance-value dependent and not just additive.

Here, we improve on the earlier approaches in the following ways: First, rather than just a generic noise term, we apply all expected nonidealities and hardware design choices (given by
\eqref{analog-mvm}) into the \ac{HWA} retraining. 
This includes dynamic range limitations, system noise, and analog-digital conversions --- all previously ignored.  
We inject weight noise in a mathematically-consistent way to both forward and backward passes, redrawing from random distributions once per mini-batch. 
We draw the weight noise from the (scaled) expected programming error distribution (see \eqref{prog_noise}) instead of using generic additive or multiplicative Gaussian distributions. 
Finally, the scale of the injected weight noise is ramped up linearly over a number of epochs, which we found to improve the \ac{HWA} training.  See \secref{experiment-methods} (Supplemental Material) for the detailed hyper-parameters and noise settings used for each \ac{DNN}.

\subsubsection*{Learning of weight-to-conductance conversion factors}
To achieve a good weight-to-conductance conversion, we train the $\affinescale_i$ scale-factors in \eqref{periphery} using \ac{SGD}. 
To improve the training of \ac{CNN} models, it is beneficial to represent these scale-factors by $\affinescale_i = \alphalearn_i \; \globalalpha$, where both the column-wise $\alphalearn_i$ and per-tile $\globalalpha$
factors can be learned. We treat the learning of either factor as a hyper-parameter for a particular \ac{DNN}. In case of not learning, $\affinescale_i$ is initialized by the weight mapping described below (see `Weight mapping and clipping`) and $\globalalpha$ is set to 1.  

In case of \acp{CNN}, where the matrix-sizes vary widely, the learned values $\alphalearn_i$ are uniquely scaled for each weight matrix by a fixed  $\alphaaws$ value, which re-scales the learning rates per tile so that the trained parameters can all have similar magnitude $\approx 1$. 
This auto-weight scaling factor, $\alphaaws$, is set to the value suggested by the Xavier weight initialization~\cite{glorot2010understanding, rasch2019training},  $\alphaaws=\sqrt{\frac{3}{n}}$, where $n$ is the input dimension of the weight matrix. 

If $\globalalpha$ is learned, we encourage the learning of larger outputs and weights by down-scaling the output range to $[-1, 1]$ which typically improves the signal-to-noise ratio, thus  $\globalalpha=\frac{\learnglobal}{\bout}$. Here $\bout$ is the fixed output bound of \eqref{periphery}, and $\learnglobal$ is a per-tile learnable scalar which is initialized to $\bout$ (and is subject to weight decay).

Note that during inference evaluation, the digital periphery can simply apply one scale-factor per output-column, since the various scale-factors described above can be re-combined after the completion of \ac{HWA} training.

\subsubsection*{Weight mapping and clipping}

Since we use the output scales $\affinescale_i$ to keep the analog weights
$\wana_{ij}$ of \eqref{analog-mvm} mapped in (normalized) conductance units
(within  $-1, \ldots, 1$), the \ac{FP} weights $w_{ij}$ of the trained
\ac{DNN} need to be mapped to conductances before initiating
\ac{HWA} training. For that we set initially
\begin{eqnarray}
  \label{eq:initial-mapping}
  \wana_{ij} &\leftarrow& \frac{w_{ij}}{\max_j|w_{ij}|}\\
  \affinescale_i &\leftarrow& \max_j|w_{ij}|
\end{eqnarray}
so that $\affinescale_i\analog{w}_{ij} = w_{ij}$.

We keep training from creating excessively large analog weights. $\wana$, by clipping after each update to this same range. 
In some cases (see Supplemental Information), we encourage learning of larger analog weights to maintain signal-to-noise ratio by remapping weights according to \eqref{initial-mapping} once every epoch. 

\subsubsection*{Learning the input range}

The input range clipping bound $c_{\text{input}}$ in \eqref{periphery} is learned during \ac{HWA} training. 
To encourage a smaller clipping value (and thus a more compact input distribution), a decay is introduced.  
To augment the gradient update for the clipping bound, we scale gradient updates by the current bound value. 
For small data sets (such as for transformer fine-tuning tasks), the \ac{HWA} training is too short to learn the clipping bound value from scratch. 
In such cases, we initialize $c_{\text{input}}$ to the average absolute maximal value of the input vectors over a number of mini-batches before starting \ac{HWA} training, subject to a cap (nominally 
$\max(c_{\text{input}}) = 10$).

\subsubsection*{Distilling with floating-point teacher}
If the model output dimension is large, such as for the \ac{LSTM} models with large vocabulary size, the \ac{HWA} training greatly benefits from distilling with the \ac{FP} model. 
In knowledge distillation~\cite{hinton2015}, an already trained ``teacher'' model augments the usual one-hot labels with expected class probabilities, which can drive a ``student'' model to a good solution more rapidly than when training only with the one-hot label vectors. 
We use the distilling  applied at the last layer, with the \ac{FP} model without any \ac{AIMC} nonidealities as the teacher and the \ac{HWA} training as described above as the student. 
The temperature controlling the distribution of pseudo-probabilities was fixed to 10, and training loss was weighted by a mixture of 75\% from the distillation and 25\% from the regular loss.

\subsection*{HWA training experiments}

We applied and optimized the \ac{HWA} training process described in this section to a variety of AI workloads -- including text prediction, speech-to-text
translation, and image classification -- as listed in \tabref{dnn-properties}. 
In general, our \ac{HWA} training approach addressed these \acp{DNN} similarly, since a too \ac{DNN}-specific re-training approach would be impractical.
In \secref{experiment-methods}, we detail any specific differences used in the \ac{HWA} training of these \acp{DNN}, including learning-rates and injected noise strength.
We select the last available rather than the best checkpoint, and we repeat experiments multiple times and average the results to obtain repeatable results.

\section{Acknowledgment}
We thank the IBM Research AI HW Center and RPI for access to the AIMOS supercomputer, and the IBM Cognitive Compute Cluster for additional compute resources. 

\bibliographystyle{abbrv}
\bibliography{refs}

\appendix
\input{supplement}

\end{document}

%% file: supplement.tex
\renewcommand\thefigure{\thesection.\arabic{figure}}
\renewcommand\thetable{\thesection.\arabic{table}}
\setcounter{figure}{0}
\setcounter{table}{0}
\newpage
\section{Supplemental Information}

\subsection{Static input range learning versus dynamic scaling}
\label{sec:input-range-learning}
In general it is difficult, if not impossible, to simultaneously achieve perfect
utilization of the input activation range, weight range, and output activation
range for every input and across every tile in the network.

In our hardware model, we assume that input ranges are statically clipped (and
scaled) by a learnable input range $\sinput$~(see
\secref{input-range-learning}). This input range is thus fixed and
re-determined during inference, and is therefore not dynamically adjusted based on
the actual values of each input vector. If dynamically adjusted, the input range
might get optimized (eg. very small inputs might not be buried in the noise
floor because of a dynamic scaling of the input range), and this could improve
the signal-to-noise ratio of the \ac{AIMC} \ac{MVM}. For instance, it was
suggested to scale the input range dynamically by the absolute maximum input
value for each input vector to maximize the output range (e.g. so-called noise
management in \cite{gokmen2017training}). One drawback of dynamic scaling,
however, is that determining the absolute max value of each input vector requires
more on-chip digital computation than simply fixing a predetermined
scaling factor. Another caveat of enlarging the input range too much is that
larger inputs might saturate the output range. In our hardware model, we assume
a relatively shallow input-to-output dynamic ratio (IO-ratio), that is we assume
10 fully-on inputs together with 10 fully-on weights (ie. at $g_\text{max}$)
would saturate the output bound so that larger output values are clipped
(compare to \figref{aimc-mvm-model}).

To test whether the trained input ranges in our static approach are sufficiently
learned, we turn on the described noise management additionally so that too
small inputs would have an improved \ac{SNR}.  To avoid a potential bound
saturation when testing the effect of the dynamic input scaling, we also turned
on the so-called bound management (which uses an dynamic down-scaling of the
input only if the output was clipped, see \cite{gokmen2017training}). We found
that across \acp{DNN}, the improvement in performance (as measured with
$\accsec$) is less than 0.1\% for all \acp{DNN} except for the transformers,
where the improvement is more pronounced but still only on the order of 1\%.
Altogether, we thus conclude that our HWA training method learned the static
input range well and could adapt the weight and input code to the shallow
IO-ratio assumption across \acp{DNN}.  A dynamic input scaling mechanism seems
not necessary to reach good accuracy for inference when using \ac{HWA} training
approach.  However, such an approach might be beneficial to increase the flexibility of models
that must be deployed with less or no \ac{HWA} training.

\begin{table}[bt]
  \centering
  {\small
  \begin{tabular}[l]{l|c|l}

    \toprule
    Parameter description & value & \textsc{AIHWKIT} configuration name \\
    \midrule
    ADC precision & 8 bit & \texttt{forward.out\_res}\\
    DAC precision & 8 bit & \texttt{forward.inp\_res}\\
    System noise referred to output & 0.04 & \texttt{forward.out\_noise}\\
    Output bound & 10.0 & \texttt{forward.out\_bound}\\
    Input bound & learned  & \texttt{forward.inp\_bound} \\
    G-max & 25$\mu$ S & \texttt{noise\_model.g\_max}\\
    (Max) tile size& 512$\times$512 & \texttt{mapping.max\_input\_size} \\ 
    Wire-conductance to $\gmax$ ratio & 571428.57 & \texttt{forward.ir\_drop\_g\_ratio} \\
    Short-term weight noise & 0.0175 & \texttt{PCM\_READ}, \texttt{forward.w\_noise}\\
    IR-drop scale & 1.0 & \texttt{forward.ir\_drop}\\
    PCM noise and drift& data-calibrated \cite{Nandakumar2019}& \texttt{PcmLikeNoiseModel}\\
    Drift compensation & global & \texttt{GlobalDriftCompensation}\\
    Layer bias & digital & \texttt{mapping.digital\_bias} \\
    Digital output scale & column-wise& \texttt{mapping.out\_scale\_columnwise}\\
    \bottomrule
  \end{tabular}
}
\caption{Parameter of our Standard PCM \ac{AIMC} model for inference (see also \figref{aimc-mvm-model-illu} and \figref{aimc-mvm-model}). The parameter names of RPU configuration settings ( \texttt{InferenceRPUConfig}) of the open-source package \ac{AIHWKIT} are indicated. Noise management and bound management techniques~\cite{gokmen2017training} are turned off and instead the learned input range is taken.  }
 \label{tab:aimc-parameter}
\end{table}

\begin{table}[tb]
  \centering



  {\small
  \begin{tabular}[l]{l|llll}
    \toprule
    Hardware-aware (HWA) training method & 1 sec & 1 hour & 1 day & 1 year \\
    \midrule
    HWA method from \cite{Joshi2020} adapted & 6.79 \std{0.01}  &  7.27 \std{0.02} & 7.80 \std{0.02} &  9.60\std{0.05}\\
    our HWA training (avg 3 training trials)  & 6.79 \std{0.02}  &  7.08 \std{0.03} & 7.50 \std{0.04} &  8.36\std{0.06}\\
    \midrule
    $\text{FP}_{32}$ mapped to analog (no HWA training) & 10.03 \std{0.18}  &  11.94 \std{0.35} & 14.27 \std{0.39} &  22.63\std{1.07}\\
    only output noise  & 7.04 \std{0.05}  &  7.40 \std{0.07} & 8.03 \std{0.09} &  10.33\std{0.25}\\
    only short-term weight noise & 7.01 \std{0.04}  &  7.62 \std{0.09} & 8.15 \std{0.12} &  10.47\std{0.38}\\
    no noise (but other nonidealities) & 7.36 \std{0.05}  &  8.04 \std{0.06} & 8.67 \std{0.18} &  11.38\std{0.25}\\
    \bottomrule
  \end{tabular}
}
\caption{\small Validation of our \ac{HWA} training methods and a number of ablations
  for a ResNet-32 model on CIFAR10~\cite{Joshi2020}. Average inference test
  errors are in percent (mean over 25 inference trials, and standard error of
  the mean). For comparison, we obtained the same weights as in
  \cite{Joshi2020}, but mapped them to our modified hardware-model
  with additional noise sources (see \eqref{periphery} and \eqref{analog-mvm})
  Our method performs results in a $>10$\% more robust model (for the most
  challenging $\tinf$) validating our approach.  Note that we here do not use
  the learning rate schedule nor the Gaussian weight clipping suggested
  by~\cite{Joshi2020}, nor any best checkpoint selection beyond hyper-parameter
  tuning (added noise strength and learning rate).  Note that it is crucial to
  add weight noise, however, as was also done in \cite{Joshi2020}.  }
 \label{tab:hwa-validation}

\end{table}

\subsection{Details on the HWA training simulations}
\label{sec:experiment-methods}

\subsubsection{LSTM on Penn Treebank Dataset}
Here the \ac{DNN} model is a two-layer \ac{LSTM} with a hidden size of 650 for
word-based prediction on the Penn Treebank Dataset (PTB)~\cite{Taylor03} using
cross entropy loss. The encoder layer is implemented digitally, whereas the
decoder layer is assumed to be on \ac{AIMC} and consists of 10K classes -- each
corresponding to a word in the dictionary. We performed \ac{HWA} training
generally as described in `AIMC hardware aware DNN training' with the following
particularities. The \ac{HWA} model is trained for $60$ epochs using a
conventionally \ac{FP} trained model as a starting point. We scan several
hyper-parameters and find the best \ac{HWA}-trained model to have a base
learning rate of $0.01$, a learning rate decay of $0.95$ (applied after each
epoch if the test error is not improving on the validation set), dropout ratio
of $0.5$, injection of \ac{PCM} programming error at $5\times$ the nominal scaling,
\ac{SGD} momentum of $0.9$, and maximum gradient norm of $10$. We use a weight
decay (L2 regularization) factor of $10^{-5}$. Finally, we randomly select 1\%
of devices to be stuck at $\gmin=0$ throughout training to provide some added
robustness (i.e. drop-connect~\cite{wan2013}).

\subsubsection{Speech-to-text LSTM with HMM}
An automatic speech recognition \ac{DNN} based on an \acf{HMM} acoustic model on
the \ac{SWB}~300 dataset \cite{cui17_interspeech} is used. The training set
consists of 262 hours of \ac{SWB}~1 audio with transcripts provided by the
Mississippi State University. The test set is the Hub5 2000 evaluation set
composed of two parts: 2.1-hour \ac{SWB} data from 40 speakers and 1.6-hour
call-home (CH) data from 40 speakers. The acoustic model is a 4-layer
bidirectional \ac{LSTM} network with input size of 140 and 512 hidden units per
direction. On top of the \ac{LSTM} layers, there is a linear projection layer
with 256 hidden units followed by a softmax output layer with 32K units
corresponding to context-dependent \ac{HMM} states. The acoustic model has a
total of 30M trainable weights.

We train the acoustic model in a \ac{HWA} manner for 20 epochs while monitoring
the validation loss, using an 8-bit \ac{FP} trained model~\cite{sun2019hybrid}
as a starting point. We use an \ac{SGD} optimizer with momentum of $0.9$, batch
size of $256$, gradient clipping of $10$, dropout of $0.1$, initial learning
rate of $0.005$ which drops by $\sqrt{10}$ at epochs $9$ and $18$, and $2\times$
scaling of the injected nominal PCM programming errors.  These were found to be the
best hyper-parameters for the \ac{HWA} trained network. We also use knowledge
distillation, which significantly improves the accuracy of the \ac{HWA}
network. In addition, we remap all weights to the full conductance range
every $2000$ mini-batches (see \eqref{initial-mapping}). The testing is
performed by pushing the input features from test-utterances through the HWA
acoustic model, simulated as being implemented in \ac{AIMC}, to get their
posteriors. Thereafter, the \ac{HMM}-based decoding network -- assumed to be implemented in
a conventional digital processor without any analog nonidealities -- is run on the
posteriors to compute the word error rate.

\subsubsection{BERT base: GLUE}
We evaluate the BERT base model~\cite{devlin2018bert} on the \ac{GLUE} benchmark, 
which nominally comprises 9 tasks. However, we excluded one task (WNLI) due to the
unusual construction of the data set and the small size of the test
set. Therefore, we evaluate 8 GLUE tasks: RTE, STS-B, MRPC, CoLA, SST-2, QNLI,
QQP, and MNLI. All linear layers of the 12~layer transformer are assumed to be deployed on  
\ac{AIMC}. In a previous study, we examined the effect of integer precision on
the attention computation~\cite{Spoon2021}.   However, here we assume that the
activation multiplications in the attention is done in \ac{FP} accuracy in
digital.  We do not train the BERT model from scratch but instead only fine-tune
the pre-trained BERT model using \ac{HWA} training techniques. We use a maximum
sequence length of $128$, because the length of the vast majority of samples are
within $128$. During fine-tuning, we scan a variety of hyper-parameters, e.g., learning rate from $1\times10^{-5}$ to $2\times10^{-4}$. We use a batch size of 5 and fixed-value weight clipping. Dropout is not used because it was not found to benefit accuracy. The number of
fine-tuning epochs varies from 5 to 20, depending on the size of the data-set for each GLUE 
task.  By using a trained \ac{FP} model as a starting point, the \ac{HWA} model
can achieve iso-accuracy with the results obtained using standard \ac{FP}
fine-tuning as a reference on most GLUE tasks. Our accuracy results are reported
on the validation data set.

\subsubsection{Albert base: GLUE}
The Albert base model is structurally identical to the BERT base model, however,
it shares the weights across the 12 layers and thus has a much smaller number of
total parameters. Similar to the BERT model, we evaluate the Albert model on the
same 8 GLUE tasks, by fine tuning the pre-trained Albert model. Similarly
hyper-parameters are scanned to identify the best parameters for the highest
accuracy, e.g., learning rate in the range from $5\times10^{-6}$ to $1.5\times10^{-4}$. Batch size is 10 and no dropout is used.

\subsubsection{ResNet-32: CIFAR10}
We used the very same ResNet-32 \ac{DNN} with slightly non-standard channel
setting as defined in the paper~\cite{Joshi2020}. Additional floating point
training was done starting from the trained model obtained via ONNX export from
the authors of~\cite{Joshi2020}, reaching slightly improved accuracy (50 batch
size; 2 steps \ac{LR} schedule with final \ac{LR} of 1\% of the starting \ac{LR}
0.025; first 2 epochs having 20 times reduced \ac{LR} warmup; 600 epochs
training with cutout augmentation; random cropping and mirroring). Using the
same \ac{LR} scheduling (but different base \ac{LR}), our standard \ac{HWA}
training was performed starting from the same ONNX model, where we assumed that
all layers are computing in \ac{AIMC}. Weights were remapped roughly once per
epoch and injected weight noise ramped up over the initial epochs.  The best
final injected programming weight noise scales was selected (from values of 3,
4, and 5). Training was repeated for 3 trials and average results are
reported. \ac{LR} was optimized among for 4 values in the range from 0.005 to
0.05.  Weight and bias decay was set to 0.001 and auto weight scaling as well as
constant down-scaling was used (see `Learning of weight-to-conductance conversion factors').

\subsubsection{WideResNet-16: CIFAR100}
We trained a WideResnet-16~\cite{zagoruyko2016wide} on the CIFAR-100
dataset~\cite{krizhevsky2009learning} using standard floating point training
(using random resize, flipping, and cutout augmentation). Then we used our
standard \ac{HWA}-training as described in `AIMC hardware aware DNN training' assuming all
layers in \ac{AIMC}. \ac{HWA} training was done for 80 epochs, with a brief
warmup and 2 step-wise learning rate reductions (to 1\% of initial
\ac{LR}). Weights were remapped roughly once per epoch~\ and injected weight
noise ramped up over the initial epochs (best of final weight noise scale of 3
and 4). We averaged the results of 3 training runs. We used auto-weight scaling,
constant down-scaling as described in `Learning of weight-to-conductance conversion factors'. Additionally,
we used bias and weight decay (0.001), and 1\% drop connect~\cite{wan2013}, and
selected the best of 6 learning rates (roughly log-spaced between 0.0005 and
0.01). Finally, we used distilling of the last activation layer with the
floating point model.

\subsubsection{ImageNet DNNs}
We used 4 standard architectures for the ImageNet data set: Resnet18 and
Resnet50~\cite{he2016deep}, WideResnet-50~\cite{zagoruyko2016wide}, and
Densenet~\cite{huang2017densely}. The main difference between the image
classification \acp{DNN} are the various layer depths and widths. As a starting
point, we used a fully trained \ac{DNN} (using floating point) that is provided
with the ``torchvision'' package~\cite{pytorch}. We then re-trained the \acp{DNN}
in \ac{HWA} manner as described in `AIMC hardware aware DNN training'. First and last layers
were assumed to be computed in digital and all others in \ac{AIMC}. The injected
weight noise scale and the learning rate were treated as a hyper-parameters. We reduced the learning rate by a factor 10 of after the initial 7-10 epochs. Batch size was adjusted to fit into the GPU memory (around 20).  Due
to computing time constraints, \ac{HWA} training was limited a few epochs (below\footnote{In case of WideResnet-50, we extended the training to 75 epochs for the last LR reduction step, which however did not change the train error significantly beyond epoch 15 due to the small learning rate.} 12) and a couple of hyper-parameter settings (e.g. noise strength
scales $1, 2, 3$ and learning rates e.g. $10^{-4}$, $5\times 10^{-5}$, $10^{-5}$ and $5\times 10^{-6}$; all combinations simulated). The best setting (typically, learning range $10^{-5}$and noise scale $3.0$) was selected based on a good compromise between short PCM drift accuracy and long-term stability. Results in \tabref{hwa-results} are presented for the very same hyper-parameter setting for all drift times. Auto-weight scaling and constant down-scaling techniques were used, as well as ramping up of the weight noise strength of the initial
epoch and occasional remapping of the affine scales during training. We also
experimented with distilling, which in some cases performed well, however, the
final results were obtained without distilling since results were found to be
more consistent.  Finally, we used a small weight decay ($10^{-5}$) and a small
weight drop connect probability of 0.1\% (\cite{wan2013}), and turned off the
IR-drop during training (but not during inference) as its effect was minimal.

\subsubsection{RNN-T speech-to-text}
The architecture consists of an acoustic encoder (transcriptor network), a
decoder (prediction network), and a joint network. The encoder comprises 6
bi-directional \ac{LSTM} layers and a \ac{FC} layer. Input size to the \ac{RNN}
is 340 (including 100-dimensional i-vectors~\cite{dehak2011ivec}), hidden size
is 640. The prediction network consists of an embedding layer (dictionary size
46, embedding vector size 10), a single unidirectional \ac{LSTM} layer with 768
cells, and a \ac{FC} layer. Encoder and prediction network outputs are combined
multiplicatively in the joint network, with a $\tanh$ activation function, and
an additional \ac{FC} layer and a final log-softmax over 46 characters.

\ac{HWA} training performs fine-tuning from a model with the same architecture,
pre-trained for 20 epochs on the audio and character-level transcripts from the
Switchboard-1 Telephone Speech Corpus (262~h of audio), further augmented with
speed and tempo perturbation \cite{ko2015audio}, SpecAugment
\cite{park2019specaugment}, and Sequence Noise Injection
\cite{saon2019sequence}. The pre-trained model achieves 11.8\% average Word
Error Rate (WER) on the Switchboard and Call-Home test sets of the NIST
Hub5-2000 evaluation.

The \ac{HWA} model is trained for few (1-5) epochs on audio sequences of
maximum length capped at 500 frames, using SGD with momentum as optimizer, with
learning rate 1e-3 and the methods described in (see
'AIMC hardware aware DNN training'). The injected weight noise scale was set to 1. Here,
neither down-scaling nor auto-scaling was used, and IR-drop was set to 0 during training.